\documentclass[runningheads]{llncs}

\usepackage{eccv}

\usepackage{eccvabbrv}

\usepackage{graphicx}
\usepackage{booktabs}

\usepackage[accsupp]{axessibility}  % Improves PDF readability for those with disabilities.

\usepackage{hyperref}

\usepackage{orcidlink}

\usepackage{svg}
\usepackage{graphicx}
\usepackage{amsmath}
\usepackage{amssymb}
\usepackage{booktabs}
\usepackage{svg}
\usepackage{booktabs}
\usepackage{algorithm}
\usepackage{algpseudocode}
\usepackage{xspace}
\usepackage{wrapfig}
\usepackage{cuted}
\usepackage{capt-of}
\usepackage{makecell}
\usepackage{enumitem}
\usepackage{xcolor}
\usepackage{colortbl} % for coloring cells
\usepackage{multirow}

\begin{document}

\title{Harnessing the Latent Diffusion Model for Training-Free Image Style Transfer} 

% \titlerunning{STRDP:}

\author{
Kento Masui\inst{1}\orcidlink{0000-0002-4174-4378} \and 
Mayu Otani\inst{1}\orcidlink{0000-0001-9923-2669} \and
Masahiro Nomura \inst{1}\orcidlink{0000-0002-4945-5984} \and
Hideki Nakayama \inst{2}\orcidlink{0000-0001-8726-2780}
}

\authorrunning{
Masui et al.
}

\institute{
CyberAgent, Japan\\
\email{\{masui\_kento,otani\_mayu,nomura\_masahiro\}@cyberagent.co.jp}
\and
The University of Tokyo, Japan\\
\email{nakayama@nlab.ci.i.u-tokyo.ac.jp}\\
}
\maketitle
\def\eg{\emph{e.g}\bmvaOneDot}
\def\Eg{\emph{E.g}\bmvaOneDot}
\newcommand{\adain}{AdaIN\xspace}
\newcommand{\ddpm}{DDPM\xspace}
\newcommand{\ldm}{LDM\xspace}

% 必要な式
\newcommand{\etheta}{\epsilon_\theta(x_t,t)}
\newcommand{\xt}{x_t}
\newcommand{\xprev}{x_{t-1}}
\newcommand{\xzero}{x_0}
\newcommand{\xzerohat}{\hat{\xzero}}
\newcommand{\zzerotoT}{z_{0:T}}
\newcommand{\zprev}{z_{t-1}}
\newcommand{\zt}{z_t}
\newcommand{\latente}{\epsilon_\theta({z_t,t})}
\newcommand{\revalphat}{\tilde{\alpha}_t}
\newcommand{\fowalphat}{\alpha_t}
\newcommand{\fowalphabar}{\bar{\alpha}}
\newcommand{\revbetat}{\tilde{\beta}_t}
\newcommand{\fowbetat}{\beta_t}
\newcommand{\fowbetabar}{\bar{\beta}}
\newcommand{\T}{\mathrm{T}}
\newcommand{\Tdash}{\T'}
\newcommand{\I}{\mathrm{I}}
\newcommand{\xlarget}{x_\mathrm{T}}
\newcommand{\Gauss}{\mathcal{N}}

\newcommand{\shape}[1]{\in \mathbb{R}^{#1}}

\newcommand{\equations}[1]{
\begin{equation}
\begin{split}
#1
\end{split}
\end{equation}
}

\newcommand{\model}{\epsilon_\theta}
\newcommand{\modelcall}{\model(\zt)}
\newcommand{\modifiedmodel}{\tilde{\epsilon}_\theta}
\newcommand{\modifiedmodelcall}{\modifiedmodel(\zt,\zst)}
\newcommand{\modelprime}{\epsilon'_{\theta}}

% Features 
\newcommand{\fstyle}{f_{sl}}
\newcommand{\fcontent}{f_{cl}}
\newcommand{\fadain}{\tilde{f}_l}

\newcommand{\xc}{x_c}
\newcommand{\xs}{x_s}
\newcommand{\xhat}{\hat{x}}
\newcommand{\zczero}{z_{c,0}}
\newcommand{\zszero}{z_{s,0}}
\newcommand{\zzero}{z_0}
\newcommand{\zct}{z_{c,t}}
\newcommand{\zst}{z_{s,t}}
\newcommand{\zcTdash}{z_{c,\mathrm{T}'}}
\newcommand{\zsTdash}{z_{s,\mathrm{T}'}} 
\newcommand{\zTdash}{z_{\mathrm{T}'}}
\newcommand{\zhatzero}{\hat{z}_0}
\newcommand{\zhatt}{\hat{z}_t}
\newcommand{\et}{\epsilon_t}

\newcommand{\hatet}{\hat{\epsilon}_t}
\newcommand{\hateprev}{\hat{\epsilon}_{t-1}}

\newcommand{\mylargefig}[3]{
\begin{figure}[t]
\centerline{\includesvg[inkscapelatex=false,width=1\columnwidth]{#1}}
\caption{#3}
\label{#2}
\end{figure}
}
\newcommand{\mylargefigsized}[4]{
\begin{figure}[t]
\centerline{\includesvg[inkscapelatex=false,width=#1\columnwidth]{#2}}
\caption{#4}
\label{#3}
\end{figure}
}

\newcommand{\mysmallfig}[3]{
\begin{wrapfigure}{T}{0.5\textwidth}
\centerline{\includesvg[inkscapelatex=false,width=0.48\textwidth]{#1}}
\caption{#3}
\label{#2}
\end{wrapfigure}
}

\newcommand{\mysmallpdf}[3]{
\begin{wrapfigure}{T}{0.5\textwidth}
\centerline{\includegraphics[width=0.5\columnwidth]{#1}}
\caption{#3}
\label{#2}
\end{wrapfigure}
}
\newcommand{\mylargepdf}[3]{
\begin{figure}[t]
\centerline{\includegraphics[width=1\columnwidth]{#1}}
\caption{#3}
\label{#2}
\end{figure}
}

\newcommand{\mylargepdfsized}[4]{
\begin{figure}[t]
\centerline{\includegraphics[width=#1\columnwidth]{#2}}
\caption{#4}
\label{#3}
\end{figure}
}

\begin{figure*}[!htb]
\centering
\includegraphics[width=1\columnwidth]{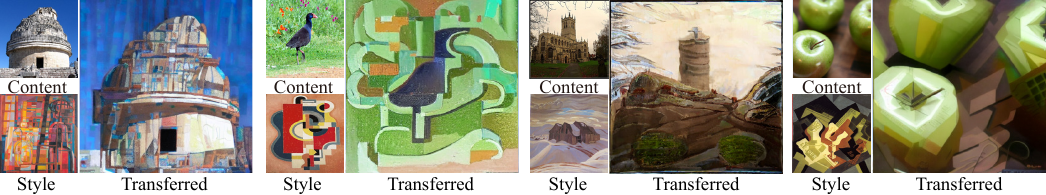}
\caption{Our image style transfer results. Our algorithm is able to transfer the visual style to a content image using a pre-trained latent diffusion model, without the need for additional training or heavy optimization. Unlike most existing approaches, our method preserves the original color of the content.}
\label{fig:teaser}
\end{figure*}

\begin{abstract}
Diffusion models have recently shown the ability to generate high-quality images. However, controlling its generation process still poses challenges. 
The image style transfer task is one of those challenges that transfers the visual attributes of a style image to another content image. 
Typical obstacle of this task is the requirement of additional training of a pre-trained model. 
We propose a training-free style transfer algorithm, Style Tracking Reverse Diffusion Process (STRDP) for a pretrained Latent Diffusion Model (LDM).
Our algorithm employs Adaptive Instance Normalization (AdaIN) function in a distinct manner during the reverse diffusion process of an LDM while tracking the encoding history of the style image. 
This algorithm enables style transfer in the latent space of LDM for reduced computational cost, and provides compatibility for various LDM models.
Through a series of experiments and a user study, we show that our method can quickly transfer the style of an image without additional training. 
The speed, compatibility, and training-free aspect of our algorithm facilitates agile experiments with combinations of styles and LDMs for extensive application.

\keywords{Image Style Transfer \and Latent Diffusion Model \and Generative Models}

\end{abstract}

\section{Introduction}
\label{sec:problem_specification}
Neural image style transfer, pioneered by Gatys \etal~\cite{gatys2016image}, is a task that transfers the visual features of a style image to another target image. In the realm of art, for instance, this technique has been used to transform photographs into painting-like images, such as those resembling Van Gogh's style. It has also been applied to represent target images as if they were composed of a given texture image. 

% TODO Add Other Related works

An essential challenge in style transfer has been its computational cost. In early research on image style transfer, Gatys \etal~demonstrated that a feature space capable of separating style and content can be obtained using VGG~\cite{simonyan2014vgg}, which was trained on a large dataset. However, their method had computational speed issues as a result of the direct optimization of images. Following Gatys, Huang \etal introduced the Adaptive Instance Normalization (AdaIN) ~\cite{huang2017arbitrary} function to approximate the style transfer effect with training a neural network. Although Huang \etal achieved a faster inference time for style transfer, training an additional neural network on top of VGG~\cite{simonyan2014vgg} still requires a considerable amount of time.

Recent generative models have made promising improvements in the speed and quality of image generation. Diffusion models~\cite{sohl-dickstein2015deep_diffusion_model,rombach2022highresolution_stablediffusion,dhariwal2021diffusion,avrahami2022blended} have enabled high-quality image generation using large-scale datasets such as LAION~\cite{schuhmann2021laion}.  In particular, a Latent Diffusion Model (LDM) known as Stable Diffusion~\cite{rombach2022highresolution_stablediffusion} has successfully learned a generative model in the latent space corresponding to the images, rather than in the image space itself. This method has improved computational speed and space by handling the diffusion process in a smaller space, making generative diffusion models feasible for consumer-grade GPUs.
%\subsubsection{Challenges}

In this paper, we propose a quick style transfer algorithm \emph{without} additional training.
Our motivation is to take advantage of the high-quality and efficient image generation capability of LDM without introducing additional training. The LDMs do not have the ability to transfer image styles; therefore, we revisit AdaIN by Huang~\etal, a function originally developed for rapid style transfer.
However, original work by Huang~\etal is not designed to work with LDM architecture and also requires additional training. Moreover, naive application of AdaIN on the LDM's latent variable does not produce desired style transfer due to its limited number of channels.
To overcome this issue, we propose an algorithm: STRDP, which alters the denoising process of LDM while iteratively applying AdaIN in a distinct repetitive manner in the U-Net architecture of LDM.
Our experiments show that our method manages color and texture styles separately, allowing styled image generation while preserving the original colors of the content image as shown in \cref{fig:teaser}

% We revisit AdaIN, a function originally developed for rapid style transfer, and adapt it to LDM that offers high-quality image generation with computational efficiency. 

%TODO AdaINを利用するにあたっての新規性を説得する
% Latent Diffusion Modelの生成品質を利用しつつ、FineTuningを伴わず、かつ軽量な手法を考える。
% LDM単体ではスタイル変換能力を持っていない。軽量な変換としてはAdaINが提案されているが、
% その設計はFineTuningを前提としている上に、LDMを想定しておらず、単純に潜在空間に適用してもチャンネル数の少なさから望む効果を得ることができない。
% この問題を解決するため、我々はLDMの逆拡散過程とU-Net architectureに着目してSTRDPを開発した
% STRDPはU-NetアーキテクチャにおいてAdaINを統合するとともに、独自のサンプリングによってスタイル変換を可能とする。

% 流れ:
% LDMの潜在空間を制御する方法は色々あるが、AdaINも検討できる。
% しかしLDMはAdaINが想定した使い方と異なるし、AdaINは学習が必要。
% 単純なアプローチとしてLDMのデコーダ向けz_0にAdaINをかけることが考えられるが、これもチャンネル数不足で効果がない
%% Decoder内でAdaINをかけても動かない
% z_0にAdaINを適用してもチャンネル数が無く動かない
% デコーダ内でAdaINしてもz_0とx_0が一対一対応しているため形状変化やスタイル変換ができない
% 計算量や学習を増やさずLDM上でスタイル変換するには、提案するアルゴリズム(Strd)が必要である。

% 学習なしQuick style transferを実現するために、我々はいくつかの選択をした
% 1. LDM特有のアーキテクチャにおいてスタイル変換を実現可能なアルゴリズムの設計
% 2. 効率的かつ高速な特徴量の転移手法としてのAdaINの選択
% 3. ２つを組み合わせた、応用性の高いアルゴリズムの実現

In summary, our contributions are as follows:

\begin{itemize}
  \item We propose an algorithm called STRDP, for a pre-trained LDM to perform style transfer without additional training. We achieve this by repeatedly applying the AdaIN function in a distinct manner in the U-Net architecture of an LDM during the reverse diffusion steps.
  \item We show that our method runs faster than other diffusion-based or training-free methods, while achieving style transfer and color preservation.
  \item We designed our algorithm to be compatible with various LDM-based models and techniques for extensive applications.
\end{itemize}

%-------------------------------------------------------------------------
% 関連研究に何があるか、問題点はなにか、この研究の立ち位置はどこか。
\section{Related Work}
\label{sec:Related_Work}
\subsection{Style Transfer Methods}
%We show two dominant approaches for style transfer methods: direct image optimization and image feature transformations.
% \paragraph{Direct Image Optimization}
% Gatys \etal~\cite{gatys2016image} proposed style loss and content loss using VGG features, realizing style transfer by directly minimizing the style and content losses between the target and style images. Gatys \etal~\cite{gatys2016preserving} also pointed out the importance of preserving color from content image by applying postprocessing algorithm such as Histogram Matching. 
% Kolkin \etal~\cite{kolkin2019style} proposed a method with loss term based on earth movers distance.  These approaches optimize image pixels using loss functions, requiring tens to hundreds of seconds to finish the task depending on the image resolution.

\paragraph{Direct Image Optimization:} % revised by GPT4
Gatys \etal~\cite{gatys2016image} introduced style and content losses using VGG features for style transfer. They also advocated for the preservation of content image through post-processing algorithms such as histogram matching~\cite{gatys2016preserving}. Kolkin \etal~\cite{kolkin2019style} employed a loss term based on Earth Mover's Distance. These methods directly optimize image pixels through loss functions, taking tens to hundreds of seconds depending on image resolution.

\paragraph{Image Feature Transformation:}
There are approaches with an encoder-decoder architecture that first encodes a content image and a style image into features~\cite{huang2017arbitrary,li2017universal,sheng2018avatar,park2019arbitrary,yoo2019photorealistic,huo2021manifold,huang2023quantart}. The features of a content image are then modified to have a style image feature. Finally, the modified feature is decoded back to an image with a style transfer effect.

Huang \etal proposed the AdaIN function to transfer the statistics of style features into the content image features. They also proposed a framework with neural network to approximate the image optimization done by Gatys. Huang \etal uses the pre-trained VGG model as an encoder. The mean and variance of the content image features in VGG are replaced with the mean and variance of the style image features using the AdaIN function. The modified features are fed to a trained decoder for the final result. Their decoder is trained to convert modified features to an image that minimizes the style and content losses proposed in \cite{gatys2016image}.

These methods are fast, as they require only a single forward pass of the model, allowing style transfer in under a second. However, they require considerable time to train the encoder-decoder architecture on which they rely.

\subsection{Diffusion Models and Controllability}
In the field of image generation, diffusion models~\cite{sohl-dickstein2015deep_diffusion_model} have achieved success in terms of generation quality with training on large datasets. 
%DDPM~\cite{ho2020denoising_ddpm} enabled learning the image generation process as long denoising steps in the image domain. 
%It allows for high-quality image generation when trained on large datasets. 
%Subsequently, DDIM~\cite{song2020denoising_ddim} significantly reduced the number of steps, enabling generation within 50 steps. 
%Furthermore, the latent diffusion model considerably improved computation space and speed by using latent space obtained by an autoencoder.
On the other hand, controlling the generation process poses a challenge. There are several approaches for controlling a trained diffusion model, as we discuss below.

\paragraph{Guidance:}
 In Guided Diffusion~\cite{dhariwal2021diffusion} and Stable Diffusion~\cite{rombach2022highresolution_stablediffusion}, the generative model is defined and learned as a conditional model. Guided Diffusion involves providing gradients from a separate model as guidance during denoising. However, both methods require training and the use of a separate model from the generative model itself, which poses a challenge. Kwon \etal~\cite{kwon2022diffuseit} also implemented style transfer in the diffusion model with guidance, but their approach is not designed to work on the latent space we aim to work on for reduced computational cost.

\paragraph{Additive Control:}
Some approaches introduce additional neural network modules or parameters into the text-to-image model to enhance controllability. 
ControlNet~\cite{Zhang2023_controlnet} incorporates a trainable clone of a diffusion model and fine-tunes new parameters with conditional input, enabling the model to generate images based on various conditions such as line arts, depth maps, etc. 
LoRA~\cite{hu2021lora} is an approach first proposed for large language models. 
LoRA introduces additional parameters into the layers of transformer architecture in a model. 
By training these parameters with different data sets and objective functions, the model can follow additional context similar to fine-tuning the original model. 
%LoRA is generic enough to be applied to text-to-image models such as Stable Diffusion.

\paragraph{Tuning:}
There are approaches that fine-tune the pre-trained LDM for controllability~\cite{ruiz2022dreambooth,kim2022diffusionclip,lu2023specialist}. InST by Zhang \etal~\cite{zhang2023inversion} also proposed a method with a text-to-image diffusion model. Their method optimizes a text embedding rather than model parameters to obtain a vector that can express the style image; however, this optimization takes 20 minutes.

We have reviewed various approaches for controlling diffusion models. However, these approaches either require additional training to enhance controllability or are not designed to work with the latent space of LDM.

\section{Background}
\paragraph{Diffusion Models:}
Diffusion Model (DM)~\cite{sohl-dickstein2015deep_diffusion_model} is a generative model that approximates the distribution of any data $x$ by first converting $x$ into simple Gaussian distribution with a diffusion process and then learning the reverse diffusion process. This diffusion process adds small amount of Gaussian noises to the original data $x_0$ over $\T$ steps until $x_T$ becomes a Gaussian noise.

Ho \etal implemented this diffusion model as Denoising Probabilistic Diffusion Model (DDPM)~\cite{ho2020denoising_ddpm} by formulating each step of the reverse diffusion process as a denoising problem. They introduced a neural network model $\etheta$ to predict the added noise of $\xt$, so that $\etheta$ can be used iteratively for $\T$ steps to reconstruct the original data $x_0$. This $\etheta$ is modeled to implicitly predict the original data $x_0$ from any $x_t$.

\paragraph{Latent Diffusion Models:}
Following the formulation of the DDPM, the Latent Diffusion Model (LDM) has been proposed by Rombach \etal~\cite{rombach2022highresolution_stablediffusion}.  LDM learns the corresponding latent space $z$ for input data $x$ using an autoencoder composed of an encoder $E$ and a decoder $D$ as $z = E(x)$ and $x = D(z)$.
LDM introduces $\latente$ to generate $z$. 
As the size of $z$ is smaller than $x$, the computational cost is significantly reduced. Note that when working with the image domain, this $\latente$ is typically modeled by a denoising U-Net~\cite{dhariwal2021diffusion,ronneberger2015u} for its ability to retain the spatial structure of the image. Unfortunately, the LDM's efficiency in latent space also introduces a challenge. The problem is that typical approaches for controlling the diffusion model, such as guidances, cannot be applied to a pre-trained LDM without additional training. 
Successive work in this field includes challenges for higher resolution image generation by SDXL~\cite{podell2023sdxl}, and faster image generation by LCM-LoRA~\cite{luo2023lcmlora}.

\paragraph{Adaptive Instance Normalization:}
To control the reverse diffusion process of a diffusion model for image style transfer, we need to modify $\latente$.
However, since $z_t$ is in the domain of latent space, statistical properties of $z_t$ must be in the valid range of the autoencoder.
Therefore, we employ AdaIN~\cite{huang2017arbitrary} function by Huang~\etal for controlling the style property of $z_t$.
AdaIN works by replacing the mean and standard deviation of each filter's activation of CNN~\cite{krizhevsky2017imagenet_cnn} for the original image with those of the style image as follows:
\begin{equation}
\text{AdaIN}(x,y) = \sigma(y)\left(\frac{x-\mu(x)}{\sigma(x)}\right) + \mu(y).
\end{equation}
$x \in \mathbb{R}^{C \times H \times W}$ is the activation of a CNN filter from the content image, and $y \in \mathbb{R}^{C \times H \times W}$ is the activation of a CNN filter from the style image. We denote $C$, $H$, and $W$ as the corresponding channel, height and width of the activations of the convolutional layer. Note that $\mu(\cdot) \in \mathbb{R}^{C}$ and $\sigma(\cdot) \in \mathbb{R}^{C}$ are channel-wise mean and standard deviation.
Huang~\etal demonstrates that the activation statistics capture visual styles for style transfer. A drawback of the framework by Huang~\etal is the requirement to train a decoder to convert the output feature into an image.

\section{Methodology}
\label{sec:Methodology}

\mylargepdf{images/TadainOverviewv17}{fig:overview}
{
An architecture of our image style transfer with style-tracking reverse diffusion process. We first add noises to the latent variables of the style and content image for $\Tdash$ steps. We keep a history of latent variables from the style image's forward diffusion process as $\zst$. In the reverse diffusion steps, we gather the CNN filter activation statistics in $\model$ from style, and transfer them to the corresponding content activations using AdaIN. This scheme allows us to transfer the image's style \emph{without} training any module. We also visualize latent variables $\zt$ and predicted noise $\hatet$ involved in this architecture as colored images. We further show a detailed diagram of $\modifiedmodel$ in \cref{fig: unet}
}

\mylargepdfsized{0.6}{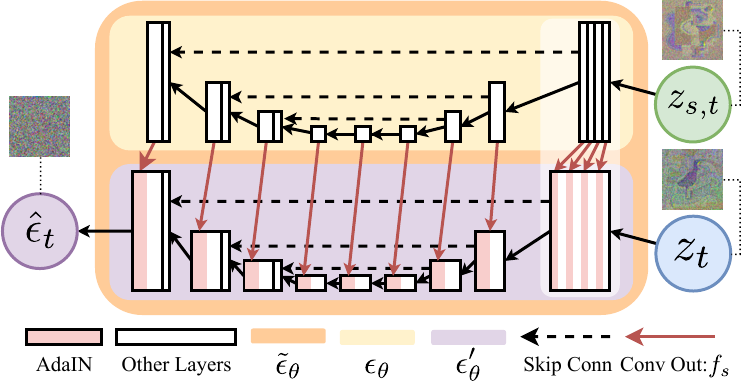}{fig: unet}{
A diagram of $\modifiedmodel$ which repeatedly applies AdaIN during the forward pass of denoising U-Net. AdaIN is introduced to every convolutional layer to transfer filter activation statistics from a style image.
}

Our key idea is to transfer visual style via representation statistics in the reverse diffusion process of pre-trained LDM using AdaIN. \Cref{fig:overview} illustrates the pipeline of our visual style transfer.

A straightforward approach to controlling the image generation process in LDM is to modulate $\zt$; however, AdaIN is not applicable to $\zt$ due to its small number of channels. We will explain how we overcome this issue in \cref{sec:reverse_diffusion}.

\subsection{Overview}
We utilize the forward and reverse diffusion processes of the LDM. First, the style image and the content image are converted to latent space. Then, in the forward diffusion process, we iteratively add noise to the representations and produce a sequence of noisy latent representations. In the reverse diffusion process, we employ AdaIN repeatedly to integrate CNN filter activations of style and content to obtain stylized latent variable. Finally, we decode the stylized latent variabele with a decoder of the LDM to obtain the stylized image.

\paragraph{Forward Diffusion Process:}

We use DDIM\cite{song2020denoising_ddim} as the base sampling algorithm, since it enables faster sampling without losing quality.
Initially, we obtain the latent space representations $\zszero \shape{C_z \times H_z \times W_z}$ and $\zczero \shape{C_z \times H_z \times W_z}$ of the style image $x_s$ and content image $x_c$ using an encoder of the LDM. Here, $C_z,H_z$, and $W_z$ denote the channel, height, and width of LDM's latent variable. 
Given the LDM scheduled to have maximum $\T$ steps for DDIM scheduling, we apply the forward diffusion process to add noise to these feature representations over $\Tdash$ steps, yielding $\zcTdash$ and $\zsTdash$. 
Here, we control $\Tdash$ with a strength parameter $S \in [0, 1]$ as $\Tdash = \textrm{round}(S*\T)$. 
During this phase, we record $\zst$ until $\zsTdash$ to use them in the decoding phase for the prediction of $\zhatzero$.
The strength $S$ controls how much noise we add to the original content and style before applying style transfer. 
If $S$ is 1, the original data becomes complete Gaussian noise in step $\Tdash = \T$, and we cannot reconstruct the original image in the reverse diffusion process. Therefore, we need to select an appropriate strength $S$ that can still retain the style and content information in $\zsTdash$ and $\zcTdash$ in step $\Tdash$.

\paragraph{Reverse Diffusion Proccess:}

Starting from the noisy latent representations $\zcTdash$ and $\zsTdash$ obtained through the forward diffusion process, we transfer the style properties from $\zst$ to $\zt$ while reconstructing the image. To derive the final latent variable $\zhatzero$ from $\zcTdash$, we track the encoding history of $\zst$. We refer to this process as the Style-Tracking Reverse Diffusion Process (STRDP). During this process, we introduce AdaIN into $\model$ to transfer the style information. After the reverse process and obtaining $\zhatzero$, the LDM decoder converts $\zhatzero$ into the final image $\xhat$.
Note that $S$ needs to be large enough to accumulate the style effect over $\Tdash$ steps in our reverse diffusion process. Therefore, $S$ controls the trade-off between the reconstruction of the content and the amount of effect applied to the result.

\subsection{Reverse Diffusion Process with AdaIN and denoising U-Net:}
\label{sec:reverse_diffusion}

To describe our reverse diffusion algorithm, we start from the following denoising equation with LDM and DDIM:
\equations{
\label{eq: reverse_step}
z_{t-1} =& \sqrt{\alpha_{t-1}}\underbrace{\left(\frac{\zt-\sqrt{1-\alpha_t}\modelcall}{\sqrt{\alpha_t}}\right)}_{\text{predicted $z_0$}} + \underbrace{\sqrt{1-\alpha_{t-1}-\sigma_t^2} \cdot \modelcall}_{\text{direction pointing to $z_t$}} + \underbrace{\sigma_t\epsilon_t}_{\text{random noise}}.
}
Here, $\alpha_t$ refers to the diffusion scheduling parameters in DDIM, while $\sigma_t$ indicates the amount of randomness introduced in the reverse diffusion process. Typically, we set $\sigma_t$ to 0 to get a deterministic result without random noise $\epsilon_t$.

\paragraph{Replacing $\model$ with $\modifiedmodel$:}
Since AdaIN for style transfer is not directly effective to $\zt$ due to its small number of channels, we apply AdaIN at CNN filters of $\model$ as they have enough number of channels to express the style~\cite{huang2017arbitrary}.
To this end, we replace the original denoising U-Net $\modelcall$ in \cref{eq: reverse_step} with our custom parallel U-Net $\modifiedmodelcall$ shown in \cref{fig: unet}.

Our parallel U-Net, $\modifiedmodelcall$, is constructed from the original U-Net $\model$ and an AdaIN-embedded $\modelprime$. We repeatedly apply the AdaIN function at every CNN filter in the model. For $l$th layer of the $\model$ and $\modelprime$, we denote $C_l, H_l, W_l$ as the layer's channel, height and width. We collect CNN feature maps $\fstyle \shape{C_l \times H_l \times W_l}$ from all convolutional layer while running $\model(\zst)$. Then, we transfer the $\mu(\fstyle) \shape{C_l}$ and $\sigma(\fstyle) \shape{C_l}$ to the corresponding feature maps $\fcontent \shape{C_l \times H_l \times W_l}$ in the AdaIN-embedded $\modelprime$. We denote the style-transferred feature maps $\fadain \shape{C_l \times H_l \times W_l}$ for each $l$th layer of $\modifiedmodel$ as follows:
\equations{
\label{eq: fadain}
\fadain =  \text{AdaIN}(\fcontent(\zt,\zst),\fstyle(\zst)),
}
which corresponds to the AdaIN layers depicted in \cref{fig: unet}.

Note that $\fadain$ depends on its incoming layer activations, which are also the result of AdaIN application, as shown in \cref{fig: unet}. This means that AdaIN is applied repeatedly during a single U-Net forward pass, in contrast to typical AdaIN application, which only applies AdaIN once to a feature. Our particular design prevents feature values from going outside the valid range in a U-Net forward pass by repeatedly applying AdaIN.

What we enforce with AdaIN is to ensure $\mu(\fadain) = \mu(\fstyle)$ and $\sigma(\fadain) = \sigma(\fstyle)$. 
This statistical constraint is conceptually similar to asking $\modifiedmodel$ to predict a noise $\hatet$ that implicitly predicts $\zhatzero$ while satisfying the constraint. In other words, $\modifiedmodel$ responds to $\zt$ as if the information from $\zst$ was present in $\zt$ regardless of its position. The nature of positional invariance of the constraint is due to AdaIN using $\mu(\cdot)$ and $\sigma(\cdot)$ for each channel, thus eliminating positional information.

\paragraph{Denoising Equation with $\modifiedmodel$:}
With our AdaIN embedded $\modifiedmodel$, we show the denoising step of our STRDP by replacing $\modelcall$ in \cref{eq: reverse_step} with $\modifiedmodelcall$ as follows:
\equations{
\label{eq: reverse_step_strd}
z_{t-1} =& \sqrt{\alpha_{t-1}}\underbrace{\left(\frac{\zt-\sqrt{1-\alpha_t}\modifiedmodelcall}{\sqrt{\alpha_t}}\right)}_{\text{predicted $z_0$}} \\
&+ \underbrace{\sqrt{1-\alpha_{t-1}-\sigma_t^2} \cdot \modifiedmodelcall}_{\text{direction pointing to $z_t$}} + \underbrace{\sigma_t\epsilon_t}_{\text{random noise}}.
}

We keep all hyperparameters the same as the original LDM implementation. We also do not modify the scheduling parameter, since the scale of our $\modifiedmodelcall$ is adjusted to be the same as $\model(\zst)$ by AdaIN function.

In summary, $\modifiedmodel$ is designed to perform style transfer by repeatedly applying AdaIN to CNN feature maps of $\modelprime$ from $\model(\zst)$ at each step of the reverse diffusion process. This particular use of AdaIN in the feature space as in \cref{eq: fadain} enables a successful style transfer without additional training. We provide a more detailed explanation of our algorithm as supplementary material, as well as the source code.

%-------------------------------------------------------------------------------------

\section{Experiments}

\label{sec:Experiments}
We compare the proposed method to prior style transfer methods, from non-diffusion based models to diffusion based models. For LDM, we used Stable Diffusion LDM-8 ~\cite{rombach2022highresolution_stablediffusion} and $\{C_z,H_z,W_z\} = \{4,64,64\}$ as the default choice.
\begin{figure}[t]
\centerline{\includegraphics[width=1\columnwidth]{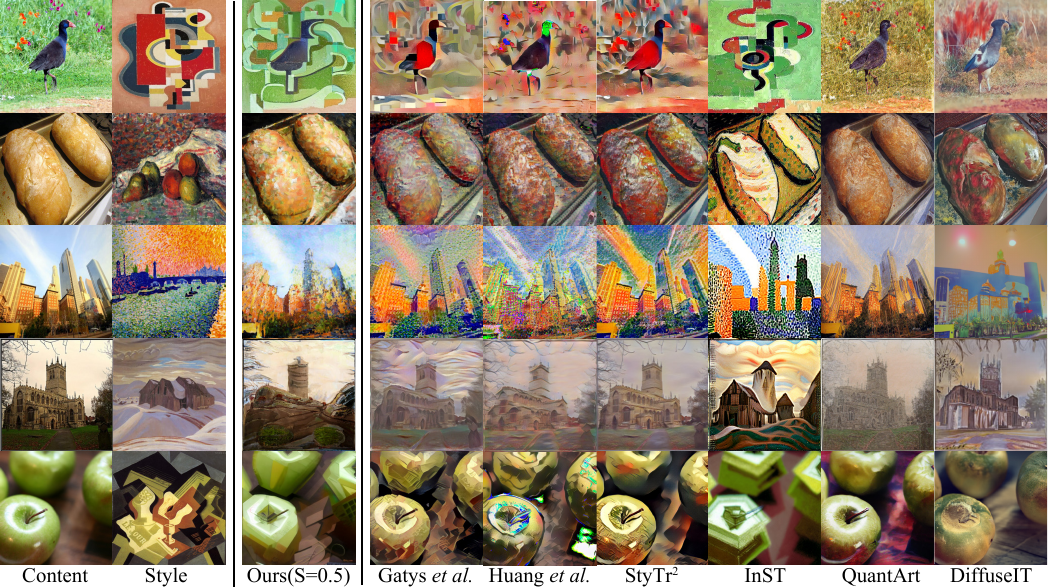}}
\caption{Qualitative comparisons of stylized images by baseline methods and ours. Our method has a texture transfer effect while preserving the color of a content image.}
\label{fig: comparison}
\end{figure}

\subsection{Comparison with State-of-the-Art Image Style Transfer}

\Cref{fig: comparison} shows examples of stylized images. 
With the strength parameter $S=0.5$, the results show that our method successfully creates stylized images.
A distinct difference is in the way these methods deal with colors.
In the top row, most prior methods involve recoloring using the style image's colors, while ours retains the original image's color tones.
This color-keeping property is beneficial for users who want to stylize an image without significantly altering its color scheme.
For Gatys~\etal and Huang~\etal, we observe severe artifacts in the output images.
InST often overlooks the content image and redraws the contents with those in the style image as seen in the third and fourth rows.
QuantArt, on the other hand, results in a conservative transformation. Although DiffuseIT transfers the style color, the results sometimes contain unrelated objects from either the style or content, such as a spotlight rendered in the third row.

\subsection{Effect of Strength $S$}
\begin{figure}[t]
\centerline{\includegraphics[width=0.8\columnwidth]{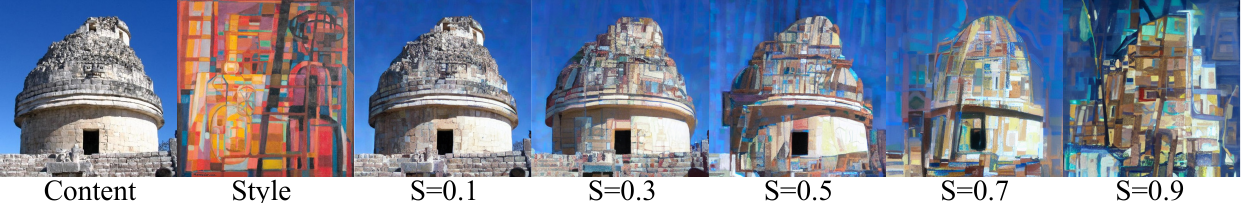}}
\caption{Visualization of the effects from $S$. The style effect becomes more apparent as we increase $S$. We can see a trade-off between the style effect and deformation. This is due to the increased reverse diffusion steps by $S$ in the LDM.}
\label{fig: strength_effect}
\end{figure}

\Cref{fig: strength_effect} visualizes the impact of strength $S \in [0, 1]$ on the output results. $S$ linearly interpolates the number of reverse diffusion steps from 0 to $\T = 50$. The results show that style transfer effect by AdaIN is strengthened as we increase $S$. At the same time, the increased number of reverse diffusion steps causes the LDM to lose the original content information, except for the color. This makes our style transfer effect a trade-off against the deformation effect.

\subsection{Computational Costs}
%\scalebox{2}{!}{%
\begin{table}[t]
    \centering
    \caption{
    Computational cost per an image, training requirement, and diffusion model choice for a 512$\times$512 pixel image. Our method consumes more memory than non-diffusion models. This is due to the base model, \ie the LDM~\cite{rombach2022highresolution_stablediffusion} is larger. However, our method does not require additional training or optimization for style control.
    }
    \begin{tabular}{l|c|c|c|c}
        \hline
        \thead{\textbf{Method}} & 
        \thead{\textbf{VRAM}} & 
        \thead{\textbf{Process Time}} & 
        \thead{\textbf{Training}} & 
        \thead{\textbf{Diffusion Model}} \\
        \hline
        \textbf{Gatys} ~\cite{gatys2016image}       & 1.3GB & 16.4 sec  & Not Required & N/A \\
        \textbf{Huang} ~\cite{huang2017arbitrary}   & 0.3GB & 0.04 sec  & Required & N/A \\
        \textbf{StyTr$^2$}~\cite{deng2022stytr2}    & 1.6GB & 0.35 sec  & Required & N/A \\
        \textbf{QuantArt} ~\cite{huang2023quantart} & 1.1GB & 0.02 sec  & Required & N/A \\
        \hline
        \textbf{InST}~\cite{zhang2023inversion}     & 7.9GB & 3.61 sec  & Required & LDM\\
        \textbf{DiffuseIT}~\cite{kwon2022diffuseit}                  & 8.6GB & 60.61 sec & Not Required & DM\\
        \hline
        \thead{\textbf{Ours} \\ {[$S=0.3$]}}                & 8.4GB & 2.70 sec & Not Required & LDM \\
        \hline
    \end{tabular}

    \label{tab:comparison}
\end{table}

% Gatys~\etal~\cite{gatys2016image}, Huang~\etal~\cite{huang2017arbitrary}, StyTr$^2$~\cite{deng2022stytr2},  QuantArt~\cite{huang2023quantart}, InST~\cite{zhang2023inversion}, and DiffuseIT~\cite{kwon2022diffuseit}.

% \cref{tab:comparison}によるベースラインとの比較によれば、我々の手法は既存手法より多くのメモリを消費するものの、処理速度は他の学習を要する手法と大差があることが確認できる。モデルの新規学習不要の手法(Gatysら）と比較すると、速度の面で提案手法に利がある。
We measured the amount of VRAM required and the computation time for the actual processing (excluding model loading).
Based on the comparison with the baselines in \cref{tab:comparison}, our method consumes more memory than the existing non-diffusion based models~\cite{gatys2016image,huang2017arbitrary,deng2022stytr2,huang2023quantart} due to the underlying LDM's size. Since we do not introduce additional parameters to the base LDM, the amount of memory required is equivalent to the LDM. The training-based methods~\cite{huang2017arbitrary,deng2022stytr2,huang2023quantart} are lightweight in inference. However, they need computationally heavy training beforehand. Compared to the existing training-free methods \cite{gatys2016image,kwon2022diffuseit}, ours has an advantage in terms of speed.

\subsection{Quantitative Evaluations}
% 提案手法とベースライン手法それぞれについて、Wikiart, Places365, ImageNetからなる100枚のランダムな組み合わせのスタイル変換結果におけるスタイル損失とコンテンツ損失を計測した。
% \begin{figure}[t]
% \centerline{\includesvg[inkscapelatex=false,width=0.5\columnwidth]{images/style_content_losses.svg}}
% \caption{The average style loss and content loss from baseline methods and ours. As the strength $S$ increases, our method produces less style loss. $S$ affects how much style information appears on the result, while content loss increases with $S$ due to increased diffusion steps.}
% \label{fig: stylecontent}
% \end{figure}

\begin{table}[t]
    \centering
        \caption{The average style loss, content loss, and quantitative metrics obtained from baseline methods and ours. As the strength $S$ increases, our method produces less style loss. $S$ affects how much style information appears on the result, while content loss increases with $S$ due to increased diffusion steps. The rest of metrics from LPIPS to CLIP consistently indicates such trade-off characteristics of our method.}

    \resizebox{0.8\columnwidth}{!}{
        \begin{tabular}{l|c|c|c|c|c|c}
\hline
 & Style Loss \textdownarrow & Content Loss \textdownarrow & LPIPS \textdownarrow & PSNR \textuparrow & SSIM \textuparrow & CLIP \textuparrow \\
 \hline
Gatys $et\ al.$ & {\cellcolor[HTML]{440154}} \color[HTML]{F1F1F1} 0.012 & {\cellcolor[HTML]{2FB47C}} \color[HTML]{F1F1F1} 16.0 & {\cellcolor[HTML]{EAE51A}} \color[HTML]{000000} 0.69 & {\cellcolor[HTML]{30698E}} \color[HTML]{F1F1F1} 16.05 & {\cellcolor[HTML]{2F6C8E}} \color[HTML]{F1F1F1} 0.40 & {\cellcolor[HTML]{38598C}} \color[HTML]{F1F1F1} 0.62 \\
Huang $et\ al.$ & {\cellcolor[HTML]{482071}} \color[HTML]{F1F1F1} 0.016 & {\cellcolor[HTML]{1FA187}} \color[HTML]{F1F1F1} 14.8 & {\cellcolor[HTML]{FDE725}} \color[HTML]{000000} 0.70 & {\cellcolor[HTML]{481769}} \color[HTML]{F1F1F1} 13.00 & {\cellcolor[HTML]{471063}} \color[HTML]{F1F1F1} 0.29 & {\cellcolor[HTML]{38598C}} \color[HTML]{F1F1F1} 0.62 \\
StyTr$^2$ & {\cellcolor[HTML]{471164}} \color[HTML]{F1F1F1} 0.014 & {\cellcolor[HTML]{25838E}} \color[HTML]{F1F1F1} 12.5 & {\cellcolor[HTML]{7AD151}} \color[HTML]{000000} 0.63 & {\cellcolor[HTML]{414487}} \color[HTML]{F1F1F1} 14.53 & {\cellcolor[HTML]{54C568}} \color[HTML]{000000} 0.54 & {\cellcolor[HTML]{20A386}} \color[HTML]{F1F1F1} 0.75 \\
QuantArt & {\cellcolor[HTML]{40BD72}} \color[HTML]{F1F1F1} 0.045 & {\cellcolor[HTML]{2E6D8E}} \color[HTML]{F1F1F1} 11.0 & {\cellcolor[HTML]{37B878}} \color[HTML]{F1F1F1} 0.58 & {\cellcolor[HTML]{3B528B}} \color[HTML]{F1F1F1} 15.08 & {\cellcolor[HTML]{3A548C}} \color[HTML]{F1F1F1} 0.37 & {\cellcolor[HTML]{40BD72}} \color[HTML]{F1F1F1} 0.80 \\
\hline
InST & {\cellcolor[HTML]{FDE725}} \color[HTML]{000000} 0.060 & {\cellcolor[HTML]{AADC32}} \color[HTML]{000000} 19.9 & {\cellcolor[HTML]{73D056}} \color[HTML]{000000} 0.62 & {\cellcolor[HTML]{414287}} \color[HTML]{F1F1F1} 14.50 & {\cellcolor[HTML]{482878}} \color[HTML]{F1F1F1} 0.31 & {\cellcolor[HTML]{481A6C}} \color[HTML]{F1F1F1} 0.54 \\
DiffuseIT & {\cellcolor[HTML]{DFE318}} \color[HTML]{000000} 0.057 & {\cellcolor[HTML]{25848E}} \color[HTML]{F1F1F1} 12.6 & {\cellcolor[HTML]{E2E418}} \color[HTML]{000000} 0.69 & {\cellcolor[HTML]{440154}} \color[HTML]{F1F1F1} 12.34 & {\cellcolor[HTML]{3C508B}} \color[HTML]{F1F1F1} 0.36 & {\cellcolor[HTML]{25858E}} \color[HTML]{F1F1F1} 0.70 \\
\hline
Ours(S=0.1) & {\cellcolor[HTML]{ADDC30}} \color[HTML]{000000} 0.054 & {\cellcolor[HTML]{440154}} \color[HTML]{F1F1F1} 4.9 & {\cellcolor[HTML]{440154}} \color[HTML]{F1F1F1} 0.33 & {\cellcolor[HTML]{FDE725}} \color[HTML]{000000} 23.28 & {\cellcolor[HTML]{FDE725}} \color[HTML]{000000} 0.63 & {\cellcolor[HTML]{FDE725}} \color[HTML]{000000} 0.93 \\
Ours(S=0.3) & {\cellcolor[HTML]{40BD72}} \color[HTML]{F1F1F1} 0.045 & {\cellcolor[HTML]{2B748E}} \color[HTML]{F1F1F1} 11.5 & {\cellcolor[HTML]{297A8E}} \color[HTML]{F1F1F1} 0.48 & {\cellcolor[HTML]{5EC962}} \color[HTML]{000000} 20.56 & {\cellcolor[HTML]{3ABA76}} \color[HTML]{F1F1F1} 0.52 & {\cellcolor[HTML]{27AD81}} \color[HTML]{F1F1F1} 0.77 \\
Ours(S=0.5) & {\cellcolor[HTML]{1F9F88}} \color[HTML]{F1F1F1} 0.038 & {\cellcolor[HTML]{3FBC73}} \color[HTML]{F1F1F1} 16.8 & {\cellcolor[HTML]{32B67A}} \color[HTML]{F1F1F1} 0.57 & {\cellcolor[HTML]{1F998A}} \color[HTML]{F1F1F1} 18.20 & {\cellcolor[HTML]{297B8E}} \color[HTML]{F1F1F1} 0.42 & {\cellcolor[HTML]{3A538B}} \color[HTML]{F1F1F1} 0.61 \\
Ours(S=0.7) & {\cellcolor[HTML]{21918C}} \color[HTML]{F1F1F1} 0.036 & {\cellcolor[HTML]{A0DA39}} \color[HTML]{000000} 19.6 & {\cellcolor[HTML]{86D549}} \color[HTML]{000000} 0.63 & {\cellcolor[HTML]{2D718E}} \color[HTML]{F1F1F1} 16.39 & {\cellcolor[HTML]{39568C}} \color[HTML]{F1F1F1} 0.37 & {\cellcolor[HTML]{48186A}} \color[HTML]{F1F1F1} 0.53 \\
Ours(S=0.9) & {\cellcolor[HTML]{23898E}} \color[HTML]{F1F1F1} 0.034 & {\cellcolor[HTML]{FDE725}} \color[HTML]{000000} 22.1 & {\cellcolor[HTML]{FDE725}} \color[HTML]{000000} 0.70 & {\cellcolor[HTML]{482979}} \color[HTML]{F1F1F1} 13.61 & {\cellcolor[HTML]{440154}} \color[HTML]{F1F1F1} 0.28 & {\cellcolor[HTML]{440154}} \color[HTML]{F1F1F1} 0.51 \\
\hline
\end{tabular}

    }

    \label{tab:quantitative}
\end{table}

To quantify the effects of style transfer, we measured the style loss and content loss used by Gatys \etal~\cite{gatys2016image} with LPIPS~\cite{zhang2018lpips}, PSNR, SSIM and CLIP~\cite{radford2021learning} similarity as supplementary. A smaller style loss indicates that the final image represents the transferred visual style. On the other hand, a small content loss indicates that the final image preserves its original content. The amount of reflected style and the preservation of content are likely to show a trade-off. We sampled 100 pairs of style and content images from WikiArt~\cite{wikiart}, ImageNet~\cite{imagenet}, and Places365~\cite{places365}
and generated stylized images using each method. The average of each metrics are shown in \Cref{tab:quantitative}.

As we increase the number of diffusion steps with $S$, we observe that the style loss decreases. This result is aligned with \cref{fig: strength_effect} since the style information becomes more prominently presented in the final results along with $S$. Our method results in larger style loss values; however, we assume that this is because our method tends to transfer style while keeping the original color of the content image. As color transfer dominates VGG-based style loss, our loss values do not decrease beyond a certain level.
For CLIP similarity, we measured cosine similarity in the CLIP encoding space. LPIPS, PSNR, SSIM, and CLIP scores indicate how content information is retained and show an aligned tendency with the content losses for our method.
In summary, we can see that $S$ controls the trade-off between the amount of style transfer effect and content retention.

\subsection{Compatibility across LDM Variants}

\mylargepdfsized{0.85}{
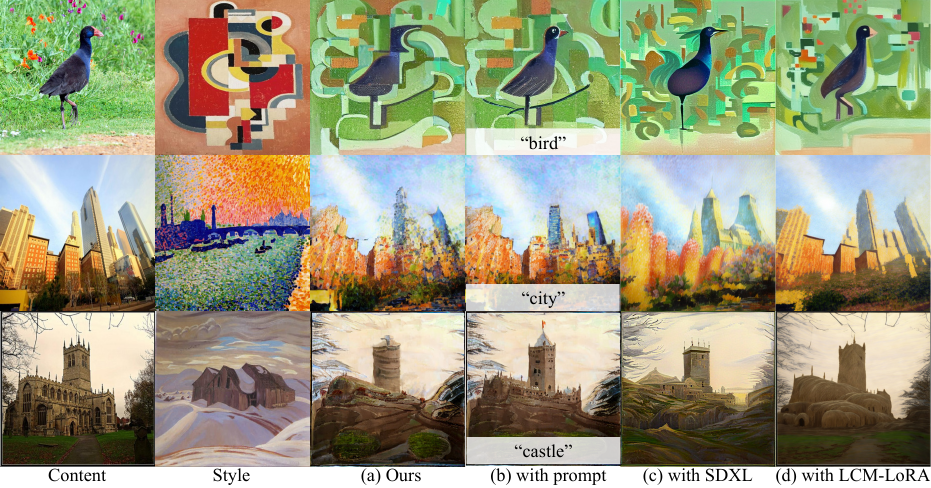
}{fig:robustness}{
Image style transfer results using our method ($S=0.5$) with text prompts and models. We can see the model follows a text prompt to maintain the features of specified object in (b), such as the bird's beak. The results with SDXL~\cite{podell2023sdxl} (c) show that our algorithm can transfer style with SDXL architecture for high resolution. The results with LCM-LoRA~\cite{luo2023lcmlora} (d) are obtained with 2 reverse steps in 0.56 seconds. This configuration (d) can achieve style transfer faster compared to standard LDM (a).
}

We experiment the compatibility of our method in \cref{fig:robustness}. We provide the results from our method with (b) text prompts, (c) SDXL~\cite{podell2023sdxl}, and (d) LCM-LoRA~\cite{luo2023lcmlora}.
SDXL is an extension of LDM to generate high-resolution images. LCM-LoRA is a combination of fine-tuning and a sampling algorithm to accelerate LDM.
For text prompts (b), we provide a word that represents the content of an image as a prompt to guide the model to generate style-transferred images while keeping their content. In (a), our method without a text prompt, the bird in the first row is strongly deformed by the abstract art style. This results in the loss of details, such as the existence of a beak. On the contrary, with a prompt as in (b), the bird has its original shape preserved even with the abstract style image. 
The results with SDXL (c) show that our algorithm can transfer style with SDXL architecture for higher image resolution ($1024\times1024$).
Our algorithm is also compatible with (d) LCM-LoRA and enables style transfer in 2 reverse steps compared to 25 reverse steps of (a). Although fine details of a style image are suppressed due to reduced reverse steps, this enables fast style transfer in less than a second.

In summary, we observe that our method can be plugged into models and techniques commonly introduced to LDM variants without training, for quick experimentations. We include the implementations for (c) and (d) in the Supplementary.

\subsection{Effect of Feature Selection and AdaIN} 
\mylargepdf{
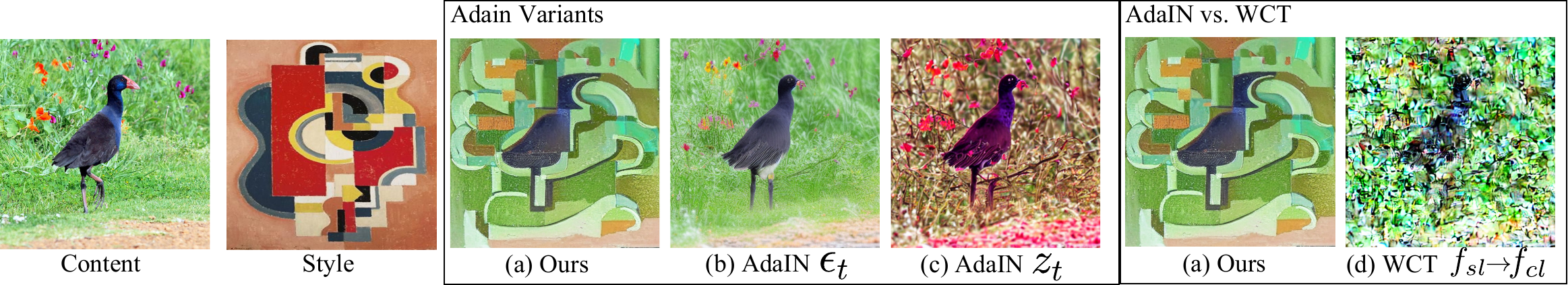
}{fig: adain_ablation}{
Ablations of applying AdaIN in various feature spaces in the reverse diffusion process with $S=0.5$. Only our approach can transfer the style effect, while other approaches that use AdaIN on different features fail to transfer styles. LDM also fails to reconstruct image from features modified by WCT.
}

To validate the design choice of AdaIN, we performed an ablation study with the selection of feature spaces to apply AdaIN. The choices of source-target pairs for AdaIN are (a): $\fstyle \rightarrow \fcontent$, \ie, ours, (b): $\model(\zst) \rightarrow \model(\zt)$, and (c): $\zst \rightarrow \zt$.
We are also interested in which feature transformation algorithms can be effectively incorporated into the diffusion model. To this end, we compared AdaIN with the whitening and coloring transform (WCT)~\cite{li2017universal} in (d).

\begin{enumerate}[label=(\alph*)]
\item $\fadain =  \text{AdaIN}(\fcontent(\zt,\zst),\fstyle(\zst))$:This proposed method repeatedly applies AdaIN in a U-Net forward pass.
\item $\hatet = \text{AdaIN}(\model(\zt),\model(\zst))$: Applying AdaIN to the predicted noise instead of CNN features in $\modifiedmodel$ as in (a).
\item $\hatet = \model(\text{AdaIN}(\zt,\zst))$: This naive approach applies AdaIN between latent variables of content and style.
\item $\tilde{f_l} = \mathrm{WCT}(\fcontent(\zt,\zst),\fstyle(\zst))$: Applying WCT instead of AdaIN for (a). WCT is also known as a method for transferring style statistics by aligning a covariance matrix between features. 
\end{enumerate}

The result is displayed in \Cref{fig: adain_ablation}.
We can see the successful style transfer effect by our method in (a).
(b) barely show style transfer effects, implying that statistics of $\model(\zst)$ do not convey the style.
(c) only transfers global color. 
Comparison between (a) ours and (d) demonstrates the effectiveness of AdaIN. (d) WCT fails to generate a stylized image. We believe this is due to WCT producing extreme values that are outside valid domains for the denoising U-Net. Moreover, introducing WCT in every CNN filter is extremely inefficient due to the computational cost of eigenvalue decomposition.
In summary, only our method (a) effectively transfers the style by applying AdaIN from $\fstyle$ to $\fcontent$, which validates our choice of design.

\subsection{Color Transfer with Histogram Matching}
\mylargepdfsized{1}{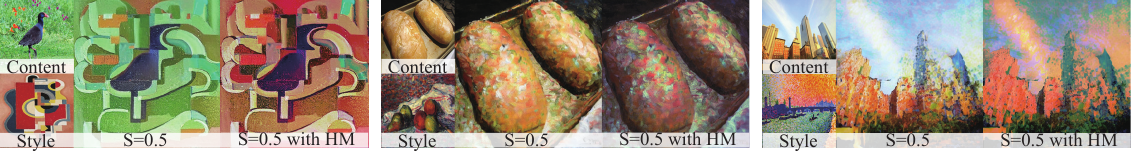}{fig: histogram_matching}{
Our style transfer results with a Histogram Matching algorithm when $S=0.5$. While our algorithm preserves original content colors, we can see that the histogram matching algorithm can additionally transfer the color style of a style image.
}

Our proposed method enables the application of texture style transfer while preserving the color of the content image. However, there may be situations where one may wish to also transfer the color distribution of the style image. While color transfer is not our primary focus, Histogram Matching can be employed as an additional post-processing step when needed. As demonstrated in \cref{fig: histogram_matching}, the use of histogram matching enables the simultaneous transfer of texture style and color distribution.

\subsection{User Study}
\label{sec: userstudy}

\begin{table}[t]
\centering
\caption{Average scores from the user study. Bold and underlined texts show the first and second best values. The scores show that our method possesses the ability to transfer texture which is comparative to Huang~\etal~\cite{huang2017arbitrary} or other diffusion based methods. Our method also achieved higher scores for color preservation.}
\resizebox{\columnwidth}{!}{
\begin{tabular}{ll|c|ccc}
    \toprule
&   &  & \multicolumn{3}{c}{Characteristics} \\
    Model & Method  & \thead{Overall Quality\textuparrow} & \thead{Color Preservation\textuparrow} & \thead{Texture Transfer\textuparrow} & \thead{Shape Preservation\textuparrow} \\
    \midrule
    \multirow{4}{*}{Non-Diffusion} & Huang~\cite{huang2017arbitrary} & 2.29 & 2.20 & 2.66 & 3.38 \\
    & Gatys~\cite{gatys2016image} & 2.89 & 2.45 & \textbf{3.65} & 3.74 \\
    & StyTr$^2$~\cite{deng2022stytr2} & \textbf{3.54} & 2.59 & \textbf{3.65} & \textbf{4.15} \\
    & QuantArt~\cite{huang2023quantart} & 3.34 & \underline{2.91} & 2.19 & \underline{4.00} \\
    \midrule
    \multirow{3}{*}{Diffusion} & InST~\cite{zhang2023inversion} & 2.17 & 2.00 & 1.92 & 1.75 \\
    & DiffuseIT~\cite{kwon2022diffuseit} & 2.59 & 2.30 & 2.44 & 2.92 \\
        \cmidrule{2-6}
    & Ours & \underline{3.39} & \textbf{4.14} & \underline{2.83} & 3.97 \\
    \bottomrule
    \end{tabular}
}
\label{table:user_study_table}
\end{table}

We conducted a user study to evaluate overall quality and 3 characteristics for image style transfer task. Characteristics include color preservation, texture transfer, and shape preservation.

In the experiment, we sampled style and content pairs from Places 365 and ImageNet dataset. We asked annotators to evaluate the output images using 5 point Likert scale, ensuring that at least three individuals answered each question. As a result, we observed 5280 responses by the annotators.
The strength $S$ of our method was fixed at 0.3. The result of the user study is in \cref{table:user_study_table}. 

In \cref{table:user_study_table}, our method achieves the second-best scores for overall quality.
Although the three characteristics do not indicate the superiority of each method, our method received a particular rating for color preservation, reflecting its tendency to retain color.
For texture transfer and shape preservation, our method is rated as having the style transfer effect while retaining the shape similar to previous methods. We provide a detailed explanation of the user study in the supplementary material.

\subsection{Limitations}
There are several limitations to our proposed method. 
First, our method is not quick enough to transfer the style effect in real time due to the base LDM. The development of faster sampling with a diffusion model such as LCM-LoRA may mitigate this issue in the future.
Second, the memory requirement for our method is comparatively larger; this is again due to the base LDM, which already consumes more memory than the models used in prior methods such as VGG. 

\section{Conclusion}
In this work, we present a style transfer algorithm called STRDP on a pretrained LDM. We introduced a custom U-Net architecture to repeatedly apply the AdaIN function in each layer during the reverse diffusion process. This approach allows quick style transfer without the need for additional training or optimization. 
We have demonstrated the style transfer performance of our algorithm in quantitative metrics and user studies, in particular, our algorithm runs faster compared to diffusion-based methods and training-free methods.
Furthermore, we have shown the compatibility of our algorithm with common LDM models and techniques, such as text prompts, SDXL for high resolution, and LCM-LoRA for faster generation. 

In summary, the speed, compatibility, and training-free characteristics of our method make it suitable for quick experimentation with image style transfer and extensive LDM-based techniques and applications.

\bibliographystyle{splncs04}
\bibliography{main}

\begin{thebibliography}{10}
\providecommand{\url}[1]{\texttt{#1}}
\providecommand{\urlprefix}{URL }
\providecommand{\doi}[1]{https://doi.org/#1}

\bibitem{wikiart}
Visual art encyclopedia (2023), \url{http://www.wikiart.org/}, accessed: 2023-05-10

\bibitem{avrahami2022blended}
Avrahami, O., Lischinski, D., Fried, O.: Blended diffusion for text-driven editing of natural images. In: Proceedings of the IEEE/CVF Conference on Computer Vision and Pattern Recognition. pp. 18208--18218 (2022)

\bibitem{imagenet}
Deng, J., Dong, W., Socher, R., Li, L.J., Li, K., Fei-Fei, L.: Imagenet: A large-scale hierarchical image database. In: 2009 IEEE Conference on Computer Vision and Pattern Recognition. pp. 248--255. IEEE (2009)

\bibitem{deng2022stytr2}
Deng, Y., Tang, F., Dong, W., Ma, C., Pan, X., Wang, L., Xu, C.: Stytr2: Image style transfer with transformers. In: Proceedings of the IEEE/CVF Conference on Computer Vision and Pattern Recognition. pp. 11326--11336 (2022)

\bibitem{dhariwal2021diffusion}
Dhariwal, P., Nichol, A.: Diffusion models beat gans on image synthesis. Advances in Neural Information Processing Systems  \textbf{34},  8780--8794 (2021)

\bibitem{gatys2016preserving}
Gatys, L., Bethge, M., Hertzmann, A., et~al.: Preserving color in neural artistic style transfer. arXiv preprint arXiv:1606.05897  (2016)

\bibitem{gatys2016image}
Gatys, L.A., Ecker, A.S., Bethge, M.: Image style transfer using convolutional neural networks. In: Proceedings of the IEEE conference on computer vision and pattern recognition. pp. 2414--2423 (2016)

\bibitem{ho2020denoising_ddpm}
Ho, J., Jain, A., Abbeel, P.: Denoising diffusion probabilistic models. Advances in Neural Information Processing Systems  \textbf{33},  6840--6851 (2020)

\bibitem{hu2021lora}
Hu, E.J., Shen, Y., Wallis, P., Allen-Zhu, Z., Li, Y., Wang, S., Wang, L., Chen, W.: Lora: Low-rank adaptation of large language models. arXiv preprint arXiv:2106.09685  (2021)

\bibitem{huang2023quantart}
Huang, S., An, J., Wei, D., Luo, J., Pfister, H.: Quantart: Quantizing image style transfer towards high visual fidelity. In: Proceedings of the IEEE/CVF Conference on Computer Vision and Pattern Recognition. pp. 5947--5956 (2023)

\bibitem{huang2017arbitrary}
Huang, X., Belongie, S.: Arbitrary style transfer in real-time with adaptive instance normalization. In: Proceedings of the IEEE international conference on computer vision. pp. 1501--1510 (2017)

\bibitem{huo2021manifold}
Huo, J., Jin, S., Li, W., Wu, J., Lai, Y.K., Shi, Y., Gao, Y.: Manifold alignment for semantically aligned style transfer. In: Proceedings of the IEEE/CVF International Conference on Computer Vision. pp. 14861--14869 (2021)

\bibitem{kim2022diffusionclip}
Kim, G., Kwon, T., Ye, J.C.: Diffusionclip: Text-guided diffusion models for robust image manipulation. In: Proceedings of the IEEE/CVF Conference on Computer Vision and Pattern Recognition. pp. 2426--2435 (2022)

\bibitem{kolkin2019style}
Kolkin, N., Salavon, J., Shakhnarovich, G.: Style transfer by relaxed optimal transport and self-similarity. In: Proceedings of the IEEE/CVF Conference on Computer Vision and Pattern Recognition. pp. 10051--10060 (2019)

\bibitem{krizhevsky2017imagenet_cnn}
Krizhevsky, A., Sutskever, I., Hinton, G.E.: Imagenet classification with deep convolutional neural networks. Communications of the ACM  \textbf{60}(6),  84--90 (2017)

\bibitem{kwon2022diffuseit}
Kwon, G., Ye, J.C.: Diffusion-based image translation using disentangled style and content representation. arXiv preprint arXiv:2209.15264  (2022)

\bibitem{li2017universal}
Li, Y., Fang, C., Yang, J., Wang, Z., Lu, X., Yang, M.H.: Universal style transfer via feature transforms. Advances in neural information processing systems  \textbf{30} (2017)

\bibitem{lu2023specialist}
Lu, H., Tunanyan, H., Wang, K., et~al.: Specialist diffusion: Plug-and-play sample-efficient fine-tuning of text-to-image diffusion models to learn any unseen style. In: Proceedings of the Conference on Computer Vision and Pattern Recognition (CVPR) (2023)

\bibitem{luo2023lcmlora}
Luo, S., Tan, Y., Patil, S., Gu, D., von Platen, P., Passos, A., Huang, L., Li, J., Zhao, H.: Lcm-lora: A universal stable-diffusion acceleration module. arXiv preprint arXiv:2311.05556  (2023)

\bibitem{park2019arbitrary}
Park, D.Y., Lee, K.H.: Arbitrary style transfer with style-attentional networks. In: proceedings of the IEEE/CVF conference on computer vision and pattern recognition. pp. 5880--5888 (2019)

\bibitem{podell2023sdxl}
Podell, D., English, Z., Lacey, K., Blattmann, A., Dockhorn, T., M{\"u}ller, J., Penna, J., Rombach, R.: Sdxl: Improving latent diffusion models for high-resolution image synthesis. arXiv preprint arXiv:2307.01952  (2023)

\bibitem{radford2021learning}
Radford, A., Kim, J.W., Hallacy, C., Ramesh, A., Goh, G., Agarwal, S., Sastry, G., Askell, A., Mishkin, P., Clark, J., et~al.: Learning transferable visual models from natural language supervision. In: International conference on machine learning. pp. 8748--8763. PMLR (2021)

\bibitem{rombach2022highresolution_stablediffusion}
Rombach, R., Blattmann, A., Lorenz, D., Esser, P., Ommer, B.: High-resolution image synthesis with latent diffusion models. In: Proceedings of the IEEE/CVF Conference on Computer Vision and Pattern Recognition. pp. 10684--10695 (2022)

\bibitem{ronneberger2015u}
Ronneberger, O., Fischer, P., Brox, T.: U-net: Convolutional networks for biomedical image segmentation. In: Medical Image Computing and Computer-Assisted Intervention--MICCAI 2015: 18th International Conference, Munich, Germany, October 5-9, 2015, Proceedings, Part III 18. pp. 234--241. Springer (2015)

\bibitem{ruiz2022dreambooth}
Ruiz, N., Li, Y., Jampani, V., Pritch, Y., Rubinstein, M., Aberman, K.: Dreambooth: Fine tuning text-to-image diffusion models for subject-driven generation. arXiv preprint arXiv:2208.12242  (2022)

\bibitem{schuhmann2021laion}
Schuhmann, C., Vencu, R., Beaumont, R., Kaczmarczyk, R., Mullis, C., Katta, A., Coombes, T., Jitsev, J., Komatsuzaki, A.: Laion-400m: Open dataset of clip-filtered 400 million image-text pairs. arXiv preprint arXiv:2111.02114  (2021)

\bibitem{sheng2018avatar}
Sheng, L., Lin, Z., Shao, J., Wang, X.: Avatar-net: Multi-scale zero-shot style transfer by feature decoration. In: Proceedings of the IEEE conference on computer vision and pattern recognition. pp. 8242--8250 (2018)

\bibitem{simonyan2014vgg}
Simonyan, K., Zisserman, A.: Very deep convolutional networks for large-scale image recognition. arXiv preprint arXiv:1409.1556  (2014)

\bibitem{sohl-dickstein2015deep_diffusion_model}
Sohl-Dickstein, J., Weiss, E., Maheswaranathan, N., Ganguli, S.: Deep unsupervised learning using nonequilibrium thermodynamics. In: International Conference on Machine Learning. pp. 2256--2265. PMLR (2015)

\bibitem{song2020denoising_ddim}
Song, J., Meng, C., Ermon, S.: Denoising diffusion implicit models. arXiv preprint arXiv:2010.02502  (2020)

\bibitem{yoo2019photorealistic}
Yoo, J., Uh, Y., Chun, S., Kang, B., Ha, J.W.: Photorealistic style transfer via wavelet transforms. In: Proceedings of the IEEE/CVF International Conference on Computer Vision. pp. 9036--9045 (2019)

\bibitem{Zhang2023_controlnet}
Zhang, L., Agrawala, M.: Adding conditional control to text-to-image diffusion models. arXiv preprint arXiv:2302.05543  (2023)

\bibitem{zhang2018lpips}
Zhang, R., Isola, P., Efros, A.A., Shechtman, E., Wang, O.: The unreasonable effectiveness of deep features as a perceptual metric. In: CVPR (2018)

\bibitem{zhang2023inversion}
Zhang, Y., Huang, N., Tang, F., Huang, H.: Inversion-based style transfer with diffusion models. In: Proceedings of the (2023)

\bibitem{places365}
Zhou, B., Lapedriza, A., Khosla, A., Oliva, A., Torralba, A.: Places: A 10 million image database for scene recognition. In: IEEE Transactions on Pattern Analysis and Machine Intelligence. vol.~40, pp. 1452--1464. IEEE (2017)

\end{thebibliography}

\end{document}

\title{Harnessing the Latent Diffusion Model for Training-Free Image Style Transfer} 

% \titlerunning{STRDP:}

\author{First Author\inst{1}\orcidlink{0000-1111-2222-3333} \and
Second Author\inst{2,3}\orcidlink{1111-2222-3333-4444} \and
Third Author\inst{3}\orcidlink{2222--3333-4444-5555}}

\authorrunning{F.~Author et al.}

\institute{Princeton University, Princeton NJ 08544, USA \and
Springer Heidelberg, Tiergartenstr.~17, 69121 Heidelberg, Germany
\email{lncs@springer.com}\\
\url{http://www.springer.com/gp/computer-science/lncs} \and
ABC Institute, Rupert-Karls-University Heidelberg, Heidelberg, Germany\\
\email{\{abc,lncs\}@uni-heidelberg.de}}

%\maketitle
\def\eg{\emph{e.g}\bmvaOneDot}
\def\Eg{\emph{E.g}\bmvaOneDot}
\newcommand{\adain}{AdaIN\xspace}
\newcommand{\ddpm}{DDPM\xspace}
\newcommand{\ldm}{LDM\xspace}

% 必要な式
\newcommand{\etheta}{\epsilon_\theta(x_t,t)}
\newcommand{\xt}{x_t}
\newcommand{\xprev}{x_{t-1}}
\newcommand{\xzero}{x_0}
\newcommand{\xzerohat}{\hat{\xzero}}
\newcommand{\zzerotoT}{z_{0:T}}
\newcommand{\zprev}{z_{t-1}}
\newcommand{\zt}{z_t}
\newcommand{\latente}{\epsilon_\theta({z_t,t})}
\newcommand{\revalphat}{\tilde{\alpha}_t}
\newcommand{\fowalphat}{\alpha_t}
\newcommand{\fowalphabar}{\bar{\alpha}}
\newcommand{\revbetat}{\tilde{\beta}_t}
\newcommand{\fowbetat}{\beta_t}
\newcommand{\fowbetabar}{\bar{\beta}}
\newcommand{\T}{\mathrm{T}}
\newcommand{\Tdash}{\T'}
\newcommand{\I}{\mathrm{I}}
\newcommand{\xlarget}{x_\mathrm{T}}
\newcommand{\Gauss}{\mathcal{N}}

\newcommand{\shape}[1]{\in \mathbb{R}^{#1}}

\newcommand{\equations}[1]{
\begin{equation}
\begin{split}
#1
\end{split}
\end{equation}
}

\newcommand{\model}{\epsilon_\theta}
\newcommand{\modelcall}{\model(\zt)}
\newcommand{\modifiedmodel}{\tilde{\epsilon}_\theta}
\newcommand{\modifiedmodelcall}{\modifiedmodel(\zt,\zst)}
\newcommand{\modelprime}{\epsilon'_{\theta}}

% Features 
\newcommand{\fstyle}{f_{sl}}
\newcommand{\fcontent}{f_{cl}}
\newcommand{\fadain}{\tilde{f}_l}

\newcommand{\xc}{x_c}
\newcommand{\xs}{x_s}
\newcommand{\xhat}{\hat{x}}
\newcommand{\zczero}{z_{c,0}}
\newcommand{\zszero}{z_{s,0}}
\newcommand{\zzero}{z_0}
\newcommand{\zct}{z_{c,t}}
\newcommand{\zst}{z_{s,t}}
\newcommand{\zcTdash}{z_{c,\mathrm{T}'}}
\newcommand{\zsTdash}{z_{s,\mathrm{T}'}} 
\newcommand{\zTdash}{z_{\mathrm{T}'}}
\newcommand{\zhatzero}{\hat{z}_0}
\newcommand{\zhatt}{\hat{z}_t}
\newcommand{\et}{\epsilon_t}

\newcommand{\hatet}{\hat{\epsilon}_t}
\newcommand{\hateprev}{\hat{\epsilon}_{t-1}}

\newcommand{\mylargefig}[3]{
\begin{figure}[t]
\centerline{\includesvg[inkscapelatex=false,width=1\columnwidth]{#1}}
\caption{#3}
\label{#2}
\end{figure}
}
\newcommand{\mylargefigsized}[4]{
\begin{figure}[t]
\centerline{\includesvg[inkscapelatex=false,width=#1\columnwidth]{#2}}
\caption{#4}
\label{#3}
\end{figure}
}

\newcommand{\mysmallfig}[3]{
\begin{wrapfigure}{T}{0.5\textwidth}
\centerline{\includesvg[inkscapelatex=false,width=0.48\textwidth]{#1}}
\caption{#3}
\label{#2}
\end{wrapfigure}
}

\newcommand{\mysmallpdf}[3]{
\begin{wrapfigure}{T}{0.5\textwidth}
\centerline{\includegraphics[width=0.5\columnwidth]{#1}}
\caption{#3}
\label{#2}
\end{wrapfigure}
}
\newcommand{\mylargepdf}[3]{
\begin{figure}[t]
\centerline{\includegraphics[width=1\columnwidth]{#1}}
\caption{#3}
\label{#2}
\end{figure}
}

\newcommand{\mylargepdfsized}[4]{
\begin{figure}[t]
\centerline{\includegraphics[width=#1\columnwidth]{#2}}
\caption{#4}
\label{#3}
\end{figure}
}

\section{Appendix}
In this appendix, we provide additional details for our methodology and experiments. We show our algorithm in \cref{sec: algorithm}, additional examples in \cref{sec: additional}, and user study details in \cref{sec: example user study}.

\section{Supplementary Video}
\label{sec: video}
We provide a short video, 'ShortPresentation.mp4', for a quick understanding of our main results and methods. 

\section{Algorithm of Style Tracking Reverse Diffusion Process}
\label{sec: algorithm}
\begin{algorithm*}
    \caption{Style Tracking Reverse Diffusion Process with Adaptive Instance Normalization and DDIM Sampling.}
    \begin{algorithmic}[1]
    \State $x_c \gets $ content image $\shape{C_x \times H_x \times W_x}$
    \State $x_s \gets $ style image $\shape{C_x \times H_x \times W_x}$
    \State $S \gets $ strength $\in (0,1)$
    \State $T \gets $ max DDIM steps 
    \State $T' \gets S \times T$ \text{// Number of forward steps to add noize into $z_{s,0}$ and $z_{c,0}$.}
    \State $z_{c,0} \gets E(x_c) \shape{C_z \times H_z \times W_z}$ 
    \State $z_{s,0} \gets E(x_s) \shape{C_z \times H_z \times W_z}$ 
    \State \text{// Calculate $z_{c,t}$ and $z_{s,t}$ with DDIM forward diffusion process for $\Tdash$ steps.}
    \For{$t \gets$ 0 to $T'$}
        \State $z_{c,t+1} \gets \mathrm{DDIM\_forward\_step}(z_{c,t},t)$
        \State $z_{s,t+1} \gets \mathrm{DDIM\_forward\_step}(z_{s,t},t)$
    \EndFor
    \State $L \gets$ number of convolution layers in $\epsilon_{\theta}$
    \State $B \gets$ batch size
    \State $C_l \gets$ number of channels in layer $l$

    \For{$l \gets 1$ to $L$}
        \State Initialize AdaIN statistics $\mu_l \in \mathbb{R}^{B \times C_l}$ and $\sigma_l \in \mathbb{R}^{B \times C_l}$ for $l$th layers.
        \State $\mu_l \gets 0$ , $\sigma_l \gets 1$
    \EndFor
    \State \text{// Denote $\mu_{all}, \sigma_{all}$ as all set of $\mu_l, \sigma_l$ for all layers.}
    \State $\mu_{all} = \{\mu_1, ..., \mu_L\}$
    \State $\sigma_{all} = \{\sigma_1, ..., \sigma_L\}$
    \State \text{// Prepare a noise prediction model $e'_{\theta}$ by replacing each convolution layer of $e_{\theta}$ to apply adain with $\mu_{all}$ and $\sigma_{all}$.}
    \State $e'_{\theta} \gets \mathrm{Clone}(e_{\theta})$
    \For{each convolution layer $l$ in $e'_{\theta}$}
        \State Replace convolution layer $l$ of $e'_{\theta}$ with $l'(x | l, \mu_l, \sigma_l) = \mathrm{AdaIN}(l(x),\mu_l,\sigma_l)$
    \EndFor
    \State Initialize $z_{T'} = z_{c,T'}$
    \For{$t \gets T'$ to $1$}
        \State \text{// Run $e_{\theta}$ with noise $z_{s_t}$ to update $\mu_{all}, \sigma_{all}$.}
        \State $\mu_{all}, \sigma_{all} \gets$ gather\_cnn\_activation\_stat($e_{\theta},z_{s_t}$)
        \State \text{// Run $e_{\theta'}$ with noise $z_t$ to get $\tilde{e}_t$ with updated $\mu_{all}, \sigma_{all}$}
        \State $\tilde{e}_t \gets e'_{\theta}(z_t | \mu_{all}, \sigma_{all})$
        \State $z_{t-1} \gets$ DDIM\_reverse\_step($z_t, \tilde{e}_t, t$)
    \EndFor
    \State \text{// Decode final $z_0$ with decoder $D$ to obtain the final image.}
    \State $\hat{x} \gets D(z_0)$
    \label{algo: strd}
    \end{algorithmic}
\end{algorithm*}
% Note that in actual python implementation, we just switch the AdaIN on and off to avoid cloning the model.
We show our algorithm of Style Tracking Reverse Diffusion Process at Algorithm 1. 
We denote the DDIM forward and reverse step functions used in Algorithm1 as

{
\small
\begin{equation}
\begin{aligned}
\mathrm{DDIM\_forward\_step}(z_{t-1},t) = \sqrt{\frac{\alpha_t}{\alpha_{t-1}}} z_{t-1} + \sqrt{1 - \alpha_t} \cdot \epsilon_t,
\end{aligned}
\end{equation}
\begin{equation}
\begin{aligned}
\mathrm{DDIM\_reverse\_step}(z_t, \hat{e}_t, t) &= \sqrt{\alpha_{t-1}} \left( \frac{z_t - \sqrt{1 - \alpha_t} \cdot \hat{e}_t}{\sqrt{\alpha_t}} \right) \\
&\quad + \sqrt{1 - \alpha_{t-1}} \cdot \hat{e}_t.
\end{aligned}
\end{equation}
}
Here, $\hat{e}_t$ refers to the predicted noise which is used for denoising in the reverse step. Note that the $\sigma_t$ is omitted since we use 0 for $\sigma_t$.

We also provide the pytorch implementation of our algorithm in the supplementary files. Upon paper publication, we will also publish a GitHub repository for our work with an implementation.

% TODO explain the DDIM forward diffusion process
% Add in Depth Explanation

\subsection{Application to Stable Diffusion XL}
Stable Diffusion XL (SDXL) consists of two U-Net architectures for the coarse image generation phase and the refinement phase. We introduced our algorithm in the SDXL refinement model. The implementation is the same as the original implementation of STRDP, except that we use the refinement model as $\model$ to create $\modifiedmodel$.

\subsection{Application to LCM-LoRA}
Latent Consistency Models (LCM)~\cite{luo2023lcm} is an algorithm to fine-tune an LDM to enable faster image generation. LCM-LoRA~\cite{luo2023lcmlora} is an extended work of LCM to obtain LoRA weights that accelerate the image generation process of an LDM.
LCM-LoRA consists of LoRA weights and a scheduler for image generation. 

The common procedure for generating an image with LCM-LoRA is as follows. 
First, a pretrained LDM is loaded.
Then an acceleration LoRA weights are applied to the loaded LDM.
Finally, a scheduler that accounts for scaling in LCM-LoRA is used to generate an image.
In the experiments for STRDP with LCM-LoRA, we followed exactly the same procedure for applying LCM-LoRA without modifying our architecture. 

We also tried directly sampling $\zst$ for each $t$ instead of iteratively adding noise to $z_{s,0}$. This resulted in obtaining the same style transfer effect despite the fructuation in $\zhatzero$ during the generation process.

\section{Additional Examples of Various Image Style Transfer Results}
\label{sec: additional}

\mylargepdf{
images/supple_comparison
}{fig:supple_comparison}{
Various style transfer results by our method with $S=0.5$. The top row shows the the content images and the left column shows the style images. We can see different types of style effects applied to each content image, depending on the style image.
}

We provide additional qualitative examples of our image style transfer results as \cref{fig:supple_comparison}.
We can see that the style effect being applied is consistent across many kind of content images when using the same style image.

\section{Examples of User Study}
\label{sec: example user study}

\begin{table}[ht]
        \centering
        \begin{tabular}{c}
        \hline
        \textbf{Inputs and Results} \\
        \hline
        \raisebox{-.5\height}{\resizebox{1\columnwidth}{!}{\includegraphics{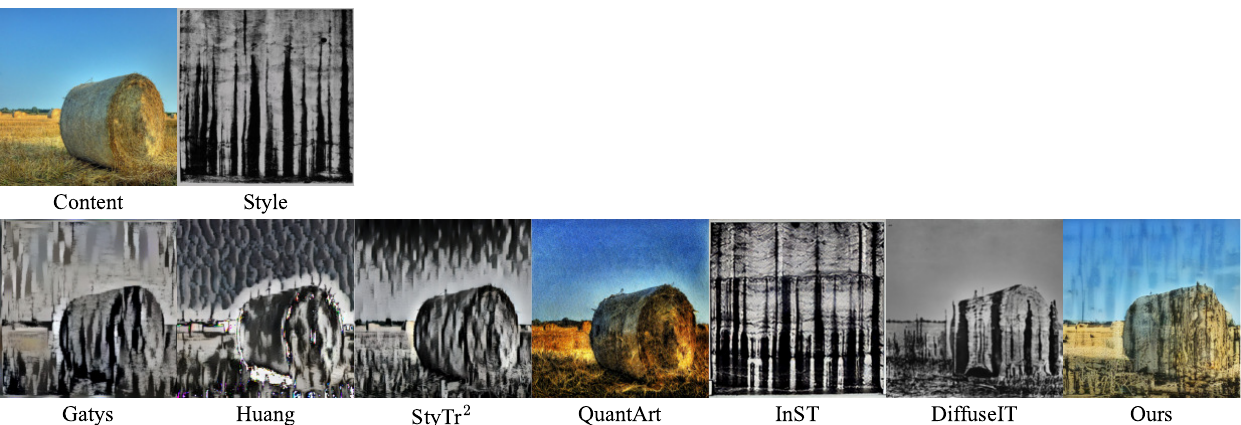}}} \\
        \begin{tabular}{l|ccccccc}
\toprule
 & Gatys $et al.$ & Huang $et al.$ & StyTr$^2$ & QuantArt & InST & DiffuseIT & Ours \\
\midrule
Overall Quality & 1.67 & 1.67 & 3.00 & \textbf{3.67} & 1.67 & 2.67 & 2.33 \\
Color Preservation & 2.33 & 2.00 & 1.00 & \textbf{3.33} & 2.33 & 1.00 & 3.00 \\
Shape Preservation & \textbf{4.00} & 3.33 & 3.67 & \textbf{4.00} & 1.67 & 3.67 & 3.33 \\
Texture Transfer & \textbf{4.00} & 2.33 & 3.33 & 1.67 & 1.33 & 3.33 & 2.67 \\
\bottomrule
\end{tabular}
 \\
        \hline
        \end{tabular}
        \caption{Examples of style transfer results and user ratings.}
        \label{tab:rating_1}
\end{table}

\begin{table}[ht]
        \centering
        \begin{tabular}{c}
        \hline
        \textbf{Inputs and Results} \\
        \hline
        \raisebox{-.5\height}{\resizebox{1\columnwidth}{!}{\includegraphics{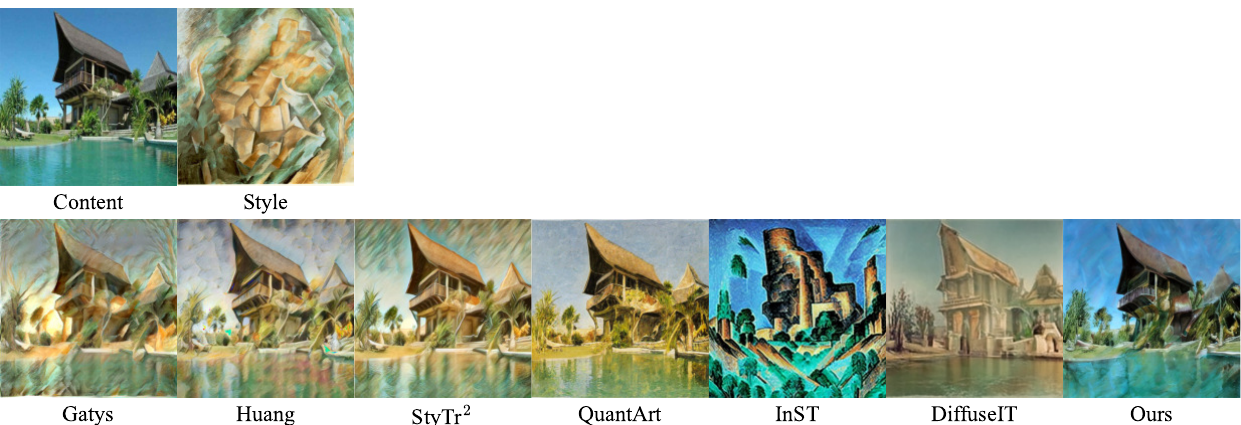}}} \\
        \begin{tabular}{l|ccccccc}
\toprule
 & Gatys $et al.$ & Huang $et al.$ & StyTr$^2$ & QuantArt & InST & DiffuseIT & Ours \\
\midrule
Overall Quality & 3.00 & 3.67 & 3.67 & \textbf{5.00} & 2.00 & 1.67 & 2.00 \\
Color Preservation & 3.33 & 3.33 & 3.33 & 3.67 & 2.33 & 3.00 & \textbf{4.67} \\
Shape Preservation & 2.67 & 3.67 & \textbf{4.33} & 3.67 & 1.67 & 2.67 & \textbf{4.33} \\
Texture Transfer & \textbf{4.67} & 3.33 & 4.33 & 3.67 & 2.67 & 2.33 & 4.00 \\
\bottomrule
\end{tabular}
 \\
        \hline
        \end{tabular}
        \caption{Examples of style transfer results and user ratings.}
        \label{tab:rating_6}
\end{table}

\begin{table}[ht]
        \centering
        \begin{tabular}{c}
        \hline
        \textbf{Inputs and Results} \\
        \hline
        \raisebox{-.5\height}{\resizebox{1\columnwidth}{!}{\includegraphics{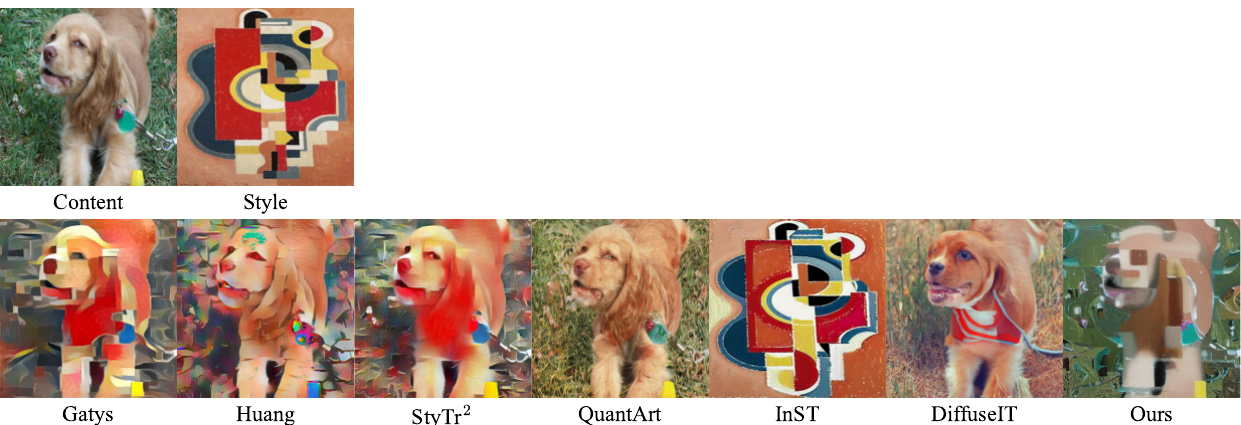}}} \\
        \begin{tabular}{l|ccccccc}
\toprule
 & Gatys $et al.$ & Huang $et al.$ & StyTr$^2$ & QuantArt & InST & DiffuseIT & Ours \\
\midrule
Overall Quality & \textbf{2.67} & 2.00 & 2.33 & 1.33 & 2.33 & 1.67 & 2.33 \\
Color Preservation & 1.67 & 2.00 & 1.33 & 2.67 & 2.33 & \textbf{3.00} & \textbf{3.00} \\
Shape Preservation & \textbf{3.67} & 3.00 & 2.67 & 3.33 & 2.00 & \textbf{3.67} & 2.33 \\
Texture Transfer & \textbf{3.33} & 2.00 & 2.33 & 1.00 & 2.33 & 1.33 & 2.00 \\
\bottomrule
\end{tabular}
 \\
        \hline
        \end{tabular}
        \caption{Examples of style transfer results and user ratings.}
        \label{tab:rating_13}
\end{table}

\begin{table}[ht]
        \centering
        \begin{tabular}{c}
        \hline
        \textbf{Inputs and Results} \\
        \hline
        \raisebox{-.5\height}{\resizebox{1\columnwidth}{!}{\includegraphics{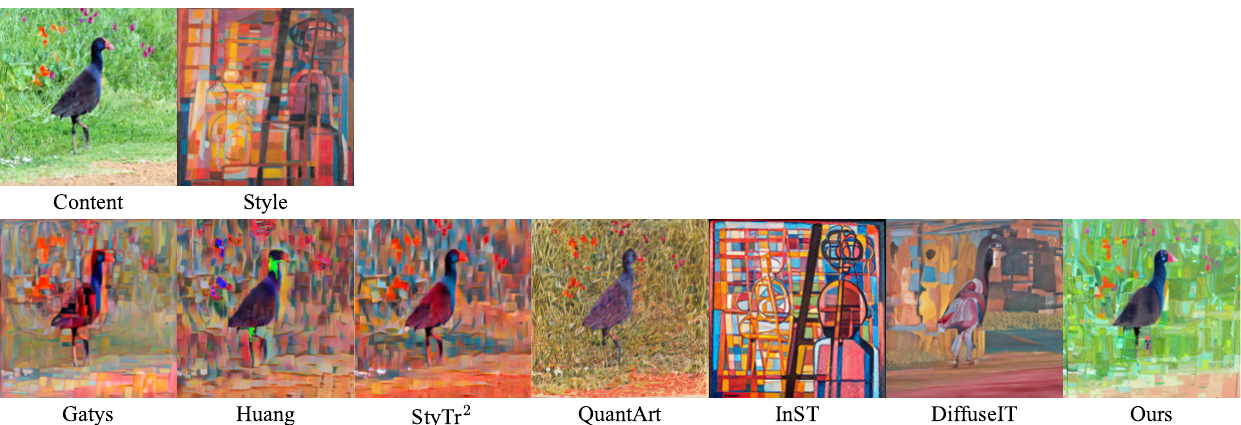}}} \\
        \begin{tabular}{l|ccccccc}
\toprule
 & Gatys $et al.$ & Huang $et al.$ & StyTr$^2$ & QuantArt & InST & DiffuseIT & Ours \\
\midrule
Overall Quality & 2.00 & \textbf{3.67} & 3.33 & 3.33 & 1.67 & 3.00 & 2.67 \\
Color Preservation & 1.67 & 2.67 & 2.67 & 2.00 & 2.33 & 2.00 & \textbf{3.00} \\
Shape Preservation & 3.00 & \textbf{3.67} & \textbf{3.67} & \textbf{3.67} & 1.00 & 3.33 & \textbf{3.67} \\
Texture Transfer & \textbf{3.33} & 2.00 & \textbf{3.33} & 2.67 & 2.00 & 1.67 & \textbf{3.33} \\
\bottomrule
\end{tabular}
 \\
        \hline
        \end{tabular}
        \caption{Examples of style transfer results and user ratings.}
        \label{tab:rating_14}
\end{table}

\begin{table}[ht]
        \centering
        \begin{tabular}{c}
        \hline
        \textbf{Inputs and Results} \\
        \hline
        \raisebox{-.5\height}{\resizebox{1\columnwidth}{!}{\includegraphics{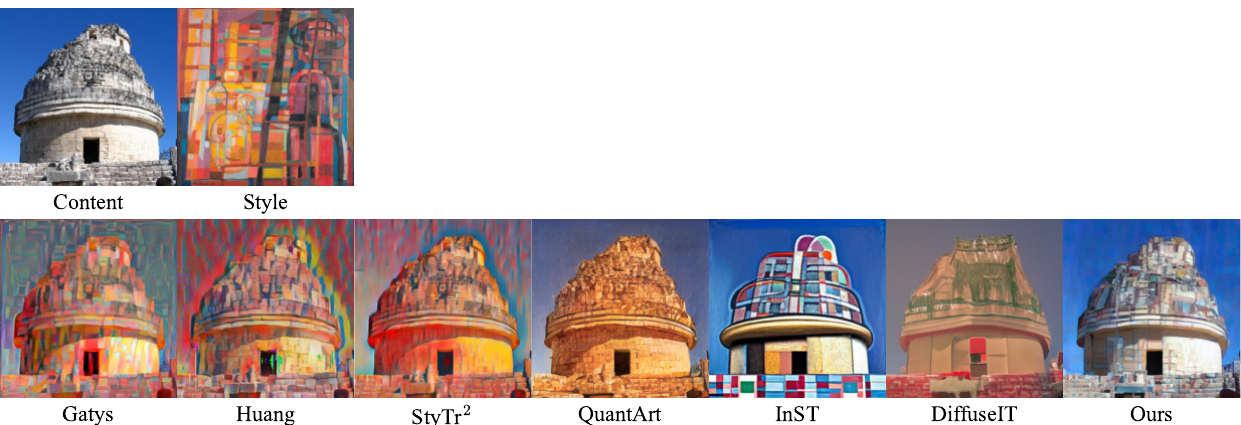}}} \\
        \begin{tabular}{l|ccccccc}
\toprule
 & Gatys $et al.$ & Huang $et al.$ & StyTr$^2$ & QuantArt & InST & DiffuseIT & Ours \\
\midrule
Overall Quality & 3.33 & 3.00 & 3.67 & 3.67 & \textbf{4.00} & 2.67 & 3.33 \\
Color Preservation & \textbf{2.67} & \textbf{2.67} & \textbf{2.67} & 1.33 & \textbf{2.67} & 1.67 & \textbf{2.67} \\
Shape Preservation & 3.33 & 4.33 & \textbf{5.00} & 3.67 & 3.67 & 2.67 & 3.67 \\
Texture Transfer & 4.00 & 3.33 & \textbf{4.67} & 1.67 & 2.67 & 1.67 & 1.67 \\
\bottomrule
\end{tabular}
 \\
        \hline
        \end{tabular}
        \caption{Examples of style transfer results and user ratings.}
        \label{tab:rating_15}
\end{table}

\begin{table}[ht]
        \centering
        \begin{tabular}{c}
        \hline
        \textbf{Inputs and Results} \\
        \hline
        \raisebox{-.5\height}{\resizebox{1\columnwidth}{!}{\includegraphics{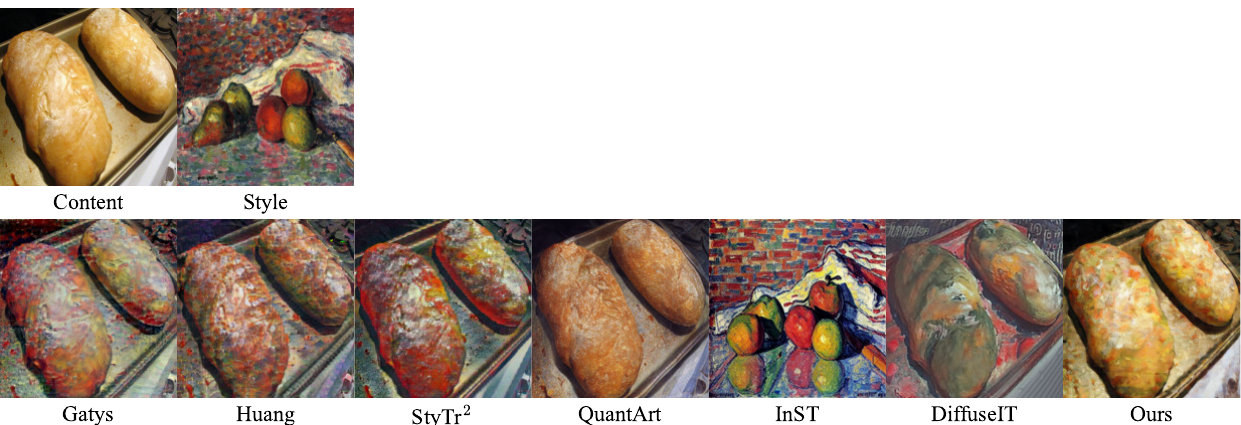}}} \\
        \begin{tabular}{l|ccccccc}
\toprule
 & Gatys $et al.$ & Huang $et al.$ & StyTr$^2$ & QuantArt & InST & DiffuseIT & Ours \\
\midrule
Overall Quality & 3.67 & 3.00 & \textbf{4.00} & 3.33 & 2.33 & 2.33 & 3.00 \\
Color Preservation & 2.67 & 2.67 & 2.67 & 3.33 & 2.00 & 1.67 & \textbf{4.00} \\
Shape Preservation & \textbf{5.00} & 4.33 & 4.67 & 4.00 & 2.67 & 2.67 & 2.33 \\
Texture Transfer & 3.67 & 3.00 & \textbf{5.00} & 2.33 & 3.00 & 3.67 & 3.33 \\
\bottomrule
\end{tabular}
 \\
        \hline
        \end{tabular}
        \caption{Examples of style transfer results and user ratings.}
        \label{tab:rating_19}
\end{table}

\begin{table}[ht]
        \centering
        \begin{tabular}{c}
        \hline
        \textbf{Inputs and Results} \\
        \hline
        \raisebox{-.5\height}{\resizebox{1\columnwidth}{!}{\includegraphics{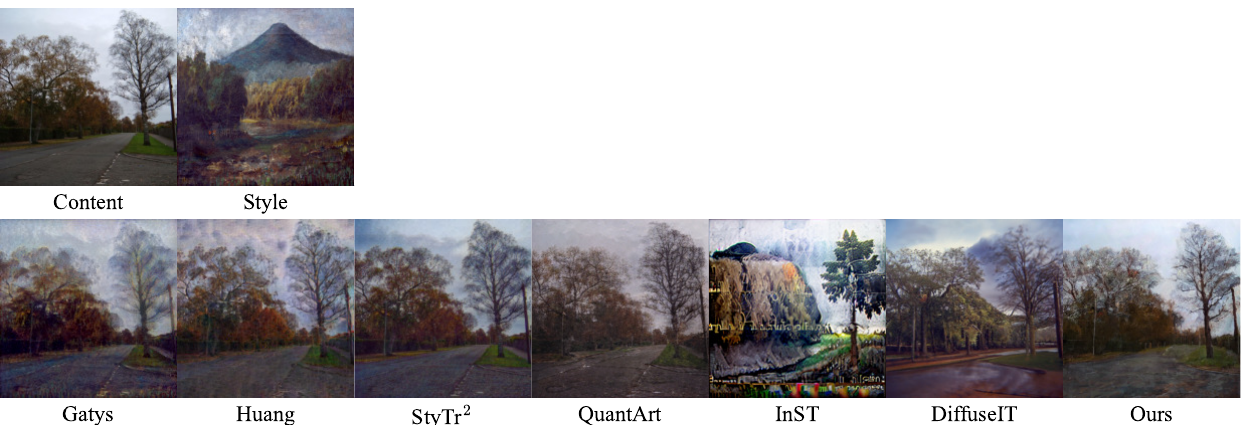}}} \\
        \begin{tabular}{l|ccccccc}
\toprule
 & Gatys $et al.$ & Huang $et al.$ & StyTr$^2$ & QuantArt & InST & DiffuseIT & Ours \\
\midrule
Overall Quality & 4.00 & 3.67 & 4.00 & \textbf{4.67} & 1.67 & 4.33 & 3.33 \\
Color Preservation & 4.33 & 3.67 & 3.33 & \textbf{4.67} & 2.00 & 4.00 & 4.00 \\
Shape Preservation & \textbf{4.67} & \textbf{4.67} & 4.33 & 4.33 & 2.00 & 4.33 & 3.67 \\
Texture Transfer & \textbf{5.00} & \textbf{5.00} & 2.67 & 3.33 & 1.67 & 3.00 & 2.00 \\
\bottomrule
\end{tabular}
 \\
        \hline
        \end{tabular}
        \caption{Examples of style transfer results and user ratings.}
        \label{tab:rating_21}
\end{table}

\begin{table}[ht]
        \centering
        \begin{tabular}{c}
        \hline
        \textbf{Inputs and Results} \\
        \hline
        \raisebox{-.5\height}{\resizebox{1\columnwidth}{!}{\includegraphics{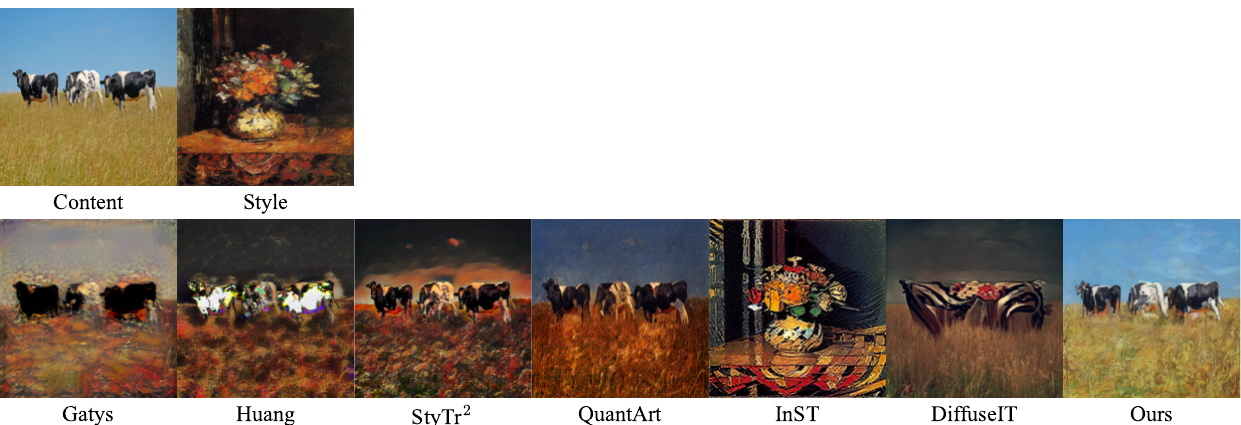}}} \\
        \begin{tabular}{l|ccccccc}
\toprule
 & Gatys $et al.$ & Huang $et al.$ & StyTr$^2$ & QuantArt & InST & DiffuseIT & Ours \\
\midrule
Overall Quality & 3.00 & 1.33 & 3.67 & \textbf{4.00} & 2.33 & 1.67 & 3.67 \\
Color Preservation & 3.00 & 1.33 & 1.33 & 3.00 & 2.33 & 1.00 & \textbf{3.33} \\
Shape Preservation & 3.67 & 2.67 & \textbf{4.67} & 4.00 & 1.67 & 2.00 & 3.00 \\
Texture Transfer & 3.33 & 2.33 & \textbf{3.67} & 2.67 & 1.67 & 1.67 & 2.33 \\
\bottomrule
\end{tabular}
 \\
        \hline
        \end{tabular}
        \caption{Examples of style transfer results and user ratings.}
        \label{tab:rating_35}
\end{table}

We show examples of the style transfer results used for our user study, along with the user ratings for each sample in \cref{tab:rating_15,tab:rating_19,tab:rating_21,tab:rating_35,tab:rating_1,tab:rating_6,tab:rating_13,tab:rating_14}.
We conducted a user study to evaluate overall quality and 3 characteristics for image style transfer task. Characteristics include color preservation, texture transfer, and shape preservation.
We used 20 style images and 20 content images to create 40 pairs of style-content images. We set the strength $S$ at 0.3. 16 annotators rated these images using a 5-point Likert scale. We made 640 questions for a mix of 4 metrics, 4 methods, and 40 style-content pairs. We made sure that at least three annotators answered each question, leading to 5280 total responses. Krippendorff's alpha\cite{krippendorff2011computing} was 0.39, indicating reliability. We investigate properties of each metrics below.

The 3 characteristics we measured are not to show the superiority of each method but to indicate the independent characteristics of each method.

\paragraph{1.~Overall Quality:}
In an overall quality assessment, an image is considered high quality when the image has no artifacts and can be used for commercial application. Our method achieves competitive overall quality. Huang~\etal tend to show some noisy artifacts compared to method by Gatys~\etal and ours. InST often overlooks the content image and generates a variant of the style image, while losing visual quality.  DiffuseIT also randomly fail to generate a meaningful image.

\paragraph{2.~Content Color Preservation:}
As we look at the content color preservation results, we observe that users evaluate our method to keep the content color the most.
This tendency is most obvious in examples where the style and content color differ largely; however, in some examples where the color of style and content are initially matched, users did not show significant differences in the results.

\paragraph{3.~Texture Transfer:}
We asked users to give a higher rating the more the texture of the style image is reproduced in the resulting image. We confirm that our method has a rating comparable to the prior methods in terms of texture transfer.

\paragraph{4.~Shape Preservation:}
We asked users to rate whether the shape of the content image remains in the final result. A higher score means that the resulting image maintains the shape of the content. Our method generally preserves the shape of the content image more than previous methods. However, abstract style images consisting of large blocks of color cause our method to deform the image larger than the previous method, as shown in the dog example in \cref{tab:rating_13}.

\bibliographystyle{splncs04}
\bibliography{main}